\documentclass[preprint,12pt]{elsarticle}

\usepackage{amsmath,amsfonts}
\usepackage{amsthm}
\usepackage{gensymb}
\usepackage{algorithmic}
\usepackage{array}
\usepackage{diagbox}
\usepackage[caption=false,font=normalsize,labelfont=sf,textfont=sf]{subfig}
\usepackage{textcomp}
\usepackage{stfloats}
\usepackage{url}
\usepackage{verbatim}
\usepackage{graphicx}
\usepackage[ruled,vlined]{algorithm2e}
\newtheorem{theorem}{Theorem}
\newtheorem{remark}{Remark}
\newtheorem{lemma}{Lemma}
\usepackage{multirow}
\usepackage{adjustbox}
\usepackage[caption=false,font=normalsize,labelfont=sf,textfont=sf]{subfig}
\usepackage{placeins}
\usepackage{soul, color}
\usepackage[utf8]{inputenc}
\usepackage[T1]{fontenc}

\journal{Digital Signal Processing}

\begin{document}

\begin{frontmatter}




\title{Downlink Channel Covariance Matrix Estimation via Representation Learning with Graph Regularization\tnoteref{title1}}
\tnotetext[label1]{This paper has been accepted by the Elsevier Digital Signal Processing and the codes of our method can be accessed via the following link: \\
\url{https://github.com/mcanzerin/downlink_ccm_estimation}}.


\author[aff1]{Melih Can Zerin\corref{cor1}\fnref{label1}}
\fntext[label1]{The work of Melih Can Zerin was supported
by Turkcell Technology under 5G and Beyond Joint Graduate
Support Program run by the Information and Communication  Technologies Authority of T\"urkiye.}
\ead{melih.zerin@metu.edu.tr}
\author[aff1]{Elif Vural}
\ead{velif@metu.edu.tr}
\author[aff1]{Ali \"Ozg\"ur Y{\i}lmaz}
\ead{aoyilmaz@metu.edu.tr}

\cortext[cor1]{Corresponding author.}

\affiliation[aff1]{organization={The Department of Electrical and Electronics Engineering, METU},
            addressline={\"{U}niversiteler Mahallesi,
Dumlup{\i}nar Bulvar{\i} No:1}, 
            city={Ankara},
            postcode={06800}, 
            country={T\"{u}rkiye}}

\begin{abstract}
In this paper, we propose an algorithm for downlink (DL) channel covariance matrix (CCM) estimation for frequency division duplexing (FDD) massive multiple-input multiple-output (MIMO) communication systems with base stations (BS) possessing a uniform linear array (ULA) antenna structure. 
We consider a setting where the UL CCM is mapped to the DL CCM by an interpolator function. We first present a theoretical error analysis of learning a nonlinear embedding by constructing an analytical mapping, which points to the importance of the Lipschitz regularity of the mapping for achieving high estimation performance. Then, based on the theoretical ground, we propose a representation learning algorithm as a solution for the estimation problem, where Gaussian RBF kernel interpolators are chosen to map UL CCMs to their DL counterparts. 
The proposed algorithm is based on the optimization of an objective function that fits a regression model between the DL CCM and the UL CCM samples in the training dataset and preserves the local geometric structure of the data in the UL CCM space, while explicitly regulating the Lipschitz continuity of the mapping function in light of our theoretical findings.
Simulation results show that the proposed algorithm surpasses benchmark methods with respect to three different error metrics. 
\end{abstract}


\begin{highlights}
\item Uplink (UL)-to-downlink (DL) channel covariance conversion is studied theoretically.
\item A representation learning algorithm is proposed based on the theoretical ground.
\item Lipschitz regularity of the embedding to be learned is shown to be important.
\item Local geometry of UL space is preserved in the DL space.
\item Proposed method outperforms benchmarks using deep learning and signal processing.
\end{highlights}


\end{frontmatter}



\section{Introduction}
\label{sec1}


Massive MIMO is a favorable technology for 5G and beyond networks in terms of achieving high spectral efficiency and reduced energy consumption \cite{bjornson2017massive}. In this technology, the base station (BS) has a much higher number of antennas than the number of active user terminals \cite{rusek2012scaling}.
The operation mode for massive MIMO is conventionally taken as time division duplexing (TDD) due to channel reciprocity \cite{bjornson2016massive}. 
Sharing the same wireless medium and frequency band, the uplink and downlink channels are said to be reciprocal in TDD systems, which means that learning the uplink channel  state information (CSI), the base station can infer the downlink CSI as well, and therefore, does not require any additional pilot training for downlink channel estimation \cite{bjornson2014massive}.
Meanwhile, the implementation of massive MIMO on FDD systems is a problem of high interest, due to the fact that most wireless networks operate in FDD mode, meaning that the infrastructure is already in place \cite{banerjee2022downlink,peng2019downlink, SADEGHI2025105287}. Additionally, FDD operation results in higher data rates and greater coverage compared to the TDD mode \cite{banerjee2022downlink,peng2019downlink}. 

\textit{Challenges of massive MIMO on FDD systems and CCM estimation.} The principle drawback of massive MIMO for FDD systems is the excessive and impractical pilot and feedback overhead \cite{bjornson2016massive,xu2014user}. The reciprocity of uplink and downlink channels does not hold for FDD systems, as they operate on different carrier frequencies  even though they share the same wireless medium  \cite{zhong2020fdd}.  As a consequence, the channel estimation process consumes too much resource for pilot and feedback symbols due to the high number of antennas at the base station \cite{bjornson2016massive}. One solution to loosen the pilot and feedback overhead is to use the DL CCM instead of the DL CSI \cite{xu2014user}. Providing the information of the second order channel statistics, the channel covariance matrix can be efficiently employed in several tasks such as channel estimation and beamforming \cite{xie2018channel,jing2024statistical}.
In many studies, the DL CCM is estimated through the UL CCM, \cite{liang2001downlink,jordan2009conversion,decurninge2015channel,khalilsarai2018fdd,miretti2021channel,banerjee2022downlink,10189874}, motivated by the
spatial reciprocity between them \cite{hugl2002spatial}
and the similarity of their power angular spectrum (PAS).

\textit{Shortcomings of existing CCM estimation methods.} While some earlier works propose simple signal processing methods for the DL CCM estimation problem, \cite{liang2001downlink,jordan2009conversion,10189874}, more recent studies have explored the utility of relatively complex models transforming the UL CCM into the DL CCM via deep learning \cite{banerjee2022downlink}. 
Methods based on signal processing may experience performance degradation in practical cases where the UL CCM is not perfectly known, as the error on the UL CCM may affect the estimate of the DL CCM.
Deep learning solutions may also be susceptible to noise in the data due to their capacity of overfitting
to a particular type of noise in the training dataset, which may eventually fail to generalize to previously unseen test points, whose noise characteristics may deviate from that of the training points.
Besides, in order to learn an accurate deep learning model with good generalization ability, one needs to use excessive amounts of data. Especially, as the number of base station antennas increases, the size of the matrix to be learned will increase along with the need for more training data.

\textit{Proposed method.} In this paper, for the DL CCM estimation problem, we propose to learn a nonlinear interpolation function which maps the UL CCM of an arbitrary user to its DL CCM. In view of the above discussions, we seek a trade-off between model simplicity and estimation performance. We thus propose to learn  a nonlinear interpolator that possesses the rich representation power of nonlinear methods with successful generalization capability, while involving a relatively small number of model parameters (e.g., much fewer than that of neural networks) to alleviate the need for training data. 
To the best of our knowledge, the estimation of DL CCMs from their UL counterparts via nonlinear interpolators has not yet been studied thoroughly in the current literature, due to which we aim to address both the theoretical and the methodological aspects of this problem.

We first present a detailed theoretical analysis, where we study the performance of mapping the UL CCM to the DL CCM via an analytical function.
We next propose an algorithm based on the minimization of the proposed objective function, which consists of a term related to the preservation of
the local neighborhood structure and two terms related to the
Lipschitz constant of the interpolator along with a data fitting term.
Our theoretical analysis shows that, under certain assumptions, the distance between two points in the DL CCM space is upper bounded proportionally to the distance between their UL CCM counterparts. This theoretical result motivates the preservation of the local neighborhood relations in the UL CCM space when mapping it to the DL CCM domain. Our theoretical analysis also indicates that
the error of an arbitrary test point decreases with decreasing values of the Lipschitz constant of the mapping.
Therefore, we also constrain the Lipschitz constant of the learnt mapping to be small in our objective function. 
We choose Gaussian RBF kernels for our interpolator, which provides a smooth interpolation of training data by preventing sudden changes in the embedding and thus avoiding overfitting, thanks to the Lipschitz regularity of the Gaussian kernel.
We use an alternating optimization method to minimize the objective function in an iterative fashion, in order to jointly learn the embedding and the parameters of the RBF interpolation function.

\textit{Key contributions.} In this paper, our main contributions to the field of DL CCM estimation from UL CCM are the following:
\begin{itemize}
    \item We first present a theoretical analysis of learning interpolation functions that map UL CCMs to their DL counterparts,
    with the purpose of identifying the main factors that affect the estimation error of the DL CCM. Our analysis shows that the error is essentially influenced by:
    (i) the average estimation error of the nearest neighbors of the point in the training dataset, (ii) the Lipschitz constant of the interpolation function, and (iii) the maximum value of the ratio of the distance between two DL CCMs to the distance between their UL counterparts.
    \item We next propose a novel representation learning method for DL CCM estimation, which builds on our theoretical results and relies on a model with much fewer parameters compared to other methods such as deep-learning algorithms. The proposed method thus achieves considerably higher estimation performance in settings with limited availability of training data. Meanwhile, the nonlinear structure of the learnt model allows for successfully capturing the particular geometry of the data, making it favorable against simpler solutions such as linear transformations.
\end{itemize}

\textit{Organization of paper.} Section \ref{related_work_section} summarizes some significant earlier works addressing the UL-DL CCM conversion problem. In Section \ref{System_Model_Section}, the system model for the communication scenario is explained. In Section \ref{Perf_Bounds_Section}, the theoretical motivation behind our method is presented. A representation learning method for the problem of DL CCM estimation from UL CCMs is proposed in Section \ref{DL_CCM_Estimation_Section}. In Section \ref{Simulation_Section}, the performance of the proposed algorithm is compared to benchmark methods via simulations in terms of several error metrics, and a stability and sensitivity analysis is presented for the proposed algorithm. Finally, the concluding remarks are given in Section \ref{Conclusion_Section}.

\textit{Notation.} A bold lower case letter such as \textbf{a} denotes a vector, while a bold upper case letter as in \textbf{A} denotes a matrix. If \textbf{A} is a square matrix, $\textbf{A}^{-1}$ and $tr(\textbf{A})$ denote the inverse and the trace of \textbf{A}, respectively. $(.)^T$ and $(.)^H$ denote the transpose and Hermitian operators, respectively.

\section{Related Work}
\label{related_work_section}
There are several efficient solutions for the estimation of the DL channel state information (CSI)  \cite{yang2019deep,utschick2022learning,sun2025hopfnet,sadeghi2023deep,fang2023sparse} and for designing feedback signals  \cite{zeng2021downlink,wang2021compressive,nerini2022machine,zhuang2024covnet}.
In \cite{adhikary2013joint}, a joint user grouping, scheduling and precoding design is developed based on CCMs of users in a multi-user environment. Similarly, \cite{sohrabi2021deep} proposes a joint pilot, feedback and precoder design in order to address the FDD massive MIMO implementation problem. In \cite{li2024downlink}, authors design an algorithm to find a pilot weighting matrix to shrink the feasible set of DL CCMs and find the center of the set in an FDD massive MIMO system with limited feedback and Type I codebook. In \cite{liu2022learning}, a neural network architecture is trained for DL CSI estimation and DL beamforming by extracting the joint long-term properties of a wireless channel that is shared by both the UL and the DL channels due to the ``partial reciprocity" of UL/DL channels.
\cite{huang2024capacity} proposes a neural network solution to optimize the achievable rate in a mmWave MIMO system with reflecting intelligent surfaces (RIS) without explicit channel estimation.

Similarly to the DL CSI estimation problem, DL CCM estimation is also a well-studied problem with a wide range of solutions available in the literature. In \cite{liang2001downlink}, the UL CCM is converted to its DL counterpart via a frequency calibration matrix that accounts for the gap between the UL and DL carrier frequencies. In \cite{jordan2009conversion}, a cubic splines method is proposed in order to interpolate the magnitude and the phase of DL CCM elements from their UL CCM counterparts. In \cite{decurninge2015channel}, a dictionary is formed from UL/DL CCM pairs, which allows the estimation of the DL CCM corresponding to an arbitrary UL CCM, by first representing it as a weighted average of the dictionary UL CCMs, and then interpolating the DL CCM from the dictionary DL CCMs with the same weights.

There are several works in the literature that explicitly exploit the angular reciprocity concept by estimating the PAS from the UL CCM and using this estimate to form the corresponding DL CCM. The methods in \cite{xie2018channel,khalilsarai2018fdd,miretti2021channel} and \cite{miretti2018fdd}  estimate the DL CCM in this manner, where the PAS is discretized for the estimation process. In \cite{xie2018channel} and \cite{khalilsarai2018fdd}, the power distribution is estimated at certain angles, which corresponds to taking discrete samples from the PAS. In \cite{miretti2021channel} and \cite{miretti2018fdd}, the UL CCM is expressed through a system of equations, from which a discrete PAS is estimated. The PAS estimation is then used to find the DL CCM of the corresponding UL CCM. 

In contrast to the above studies, using the UL CCM  to directly estimate the DL CCM without explicitly finding the PAS is also an option, which is addressed in several works such as \cite{liang2001downlink,decurninge2015channel,banerjee2022downlink,10189874}. 
The method in \cite{decurninge2015channel} employs a dictionary of UL/DL CCM pairs for the deduction of the DL CCM of a new user with the help of its UL CCM and the dictionary. The study in \cite{banerjee2022downlink} adapts the image-to-image translation idea in  \cite{isola2017image} to CCM estimation by converting CCMs into RGB images and processing them via a conditional generative adversarial network (CGAN) architecture.
A variational autoencoder is used in \cite{linfu2021deep} for translating UL CCMs to their DL counterparts, by representing UL CCMs and DL CCMs as images, similarly to \cite{banerjee2022downlink}.
In \cite{10189874}, UL CCM entries are considered to be related to the common PAS of the UL and DL channels through a nonlinear transformation. Based on this model, a linear transformation that maps an UL CCM to its DL CCM is proposed. Our study bears resemblance to these aforementioned methods in that it also presents a direct estimation approach without explicitly computing the PAS. In literature, machine learning algorithms that address the UL-to-DL CCM mapping problem generally use deep neural networks for this task. Although deep learning methods are able to learn highly complex models, they require tremendous amounts of data for successful generalization, in contrast to simpler nonlinear interpolator structures with fewer parameters as chosen in our work.

In \cite{ornek2019nonlinear}, the performance of learning a supervised nonlinear embedding via a mapping function is examined for classification problems, where particular attention is paid to the generalization of the learned embedding to previously unseen data. While we leverage results from \cite{ornek2019nonlinear} to address some of the technicalities of the proofs in this paper, the aim and the scope of the current study are essentially different from those of \cite{ornek2019nonlinear}, as we develop a regression framework specifically designed for solving a wireless communications problem, i.e., UL-to-DL CCM transformation. Our theoretical analysis provides performance bounds for this particular problem setting.

\section{System Model}
\label{System_Model_Section}
We consider an FDD single cell massive MIMO system, in which a base station (BS) containing $M$ antennas forming a uniform linear array (ULA) serves single-antenna user equipments (UE). The UL channel operates at the carrier frequency $f_{UL}$ and the DL channel operates at the carrier frequency $f_{DL}$, with respective wavelengths  $\lambda_{UL}$ and $\lambda_{DL}$. We denote the ratio of carrier frequencies as $f_{R}=\frac{f_{DL}}{f_{UL}}=\frac{\lambda_{UL}}{\lambda_{DL}}$. The UL and the DL channels are considered to be frequency-flat.

The UL and the DL channel vectors ($\textbf{h}_{UL}$ and $\textbf{h}_{DL}$, respectively) are modeled as  \cite{banerjee2022downlink}, \cite{10189874}

\begin{equation}
    \textbf{h}_x=\int_{\bar{v}-\Delta}^{\bar{v}+\Delta} \gamma_x(\phi)\textbf{a}_{x} (\phi) \,d\phi, x \in \{{UL,DL}\},
\end{equation}
where $\gamma_x(\phi)$ is the complex channel gain corresponding to the angle of arrival (AoA) $\phi$ and $\textbf{a}_{x} (\phi)$ is the array response vector at the angle $\phi$.
The array response vectors of the UL and the DL channels ($\textbf{a}_{UL} (\phi)$ and $\textbf{a}_{DL} (\phi)$, respectively) are given by
\begin{equation} \label{array_response_vector}
    \textbf{a}_{x} (\phi) = [1 \ e^{j2\pi \frac{d}{\lambda_{x}} \sin{\phi}} \  ... \ e^{j2\pi \frac{d}{\lambda_{x}} (M-1) \sin{\phi}} ] ^{T},
    x \in \{{UL,DL}\},
\end{equation}
where $d = \frac{\lambda_{UL}}{2}$ is the distance between the adjacent antenna elements at the BS. 

We consider the wide sense stationary uncorrelated scattering (WSSUS) model for our communication scenario as in \cite{10189874}. In this model, the autocorrelation function (acf) of the channel gain is time-invariant, and the scattering at different AoA's is uncorrelated. 
Considering the UL and the DL channels as zero-mean, the UL CCM and the DL CCM ($\textbf{R}_{UL}$ and $\textbf{R}_{DL}$, respectively) can then be formulated as \cite{10189874}

\begin{multline} \label{CCM_Formula}
    \textbf{R}_{x} = \mathop{\mathbb{E}} \left\{\left(\textbf{h}_{x}-\mathop{\mathbb{E}} \left\{ \textbf{h}_{x}\right\}  \right) \left(\textbf{h}_{x}-\mathop{\mathbb{E}} \left\{ \textbf{h}_{x}\right\}  \right)^{H} \right\}= \mathop{\mathbb{E}} \left\{\textbf{h}_{x} \textbf{h}_{x}^{H} \right\} \\ =  \int_{\bar{v}-\Delta}^{\bar{v}+\Delta} p(\phi)\textbf{a}_{x} (\phi) \textbf{a}_{x}^{H} (\phi) \,d\phi, x \in \{{UL,DL}\},
\end{multline}
where  $p(\phi)$ is the power angular spectrum (PAS), $\Delta$ is the spread of AoAs and $\bar{v}$ is the mean AoA. The PAS is the same for uplink and downlink and normalized to 1, i.e., $\int_{\bar{v}-\Delta}^{\bar{v}+\Delta} p(\phi) d\phi =1$. From \eqref{CCM_Formula} and \eqref{array_response_vector}, one can conclude that CCMs are Hermitian, i.e., $\textbf{R}_x=\textbf{R}_x^{H}$, for $x \in \{{UL,DL}\}$.
The ULA antenna structure and the WSSUS model cause the CCMs to be Toeplitz.
Due to its Hermitian and Toeplitz structure, the $\textbf{R}_x$ matrix given in \eqref{CCM_Formula} is fully characterized by its first row.

\section{Performance Bounds for DL CCM Estimation via Gaussian RBF kernels} \label{Perf_Bounds_Section}

In this section, we first present the representation learning setting proposed for DL CCM estimation from UL CCM. We then provide an upper bound on the error of an arbitrary test sample.

\subsection{Notation and Setting}
Let $\{{\textbf{r}_{UL}}^{i},{\textbf{r}_{DL}}^{i}\}_{i=1}^{N}$ be a training dataset with $N$ training UL/DL CCM sample pairs, where ${\textbf{r}_{x}}^{i} \in \mathbb{R}^{1 \times 2M-1} $ is a row vector obtained by the concatenation of the real and imaginary parts of the first row vector of the $i^{th}$ CCM in the training dataset for $x \in \{ UL, DL\}$. The first element of the first row of a CCM is always real with  no imaginary part. Hence,  the vectors in the dataset are of length $2M-1$. Let the UL data samples be drawn i.i.d. from a probability measure $\upsilon$ on $\mathbb{R}^{1 \times 2M-1}$. The training samples are embedded into $\mathbb{R}^{1 \times 2M-1}$ such that each training sample ${\textbf{r}_{UL}}^{i}$ is mapped to a vector ${{\hat{\textbf{r}}}_{DL}}^{i}\in\mathbb{R}^{1 \times 2M-1}$.The mapping is assumed to be extended to the whole data space through an interpolation function $f:\mathbb{R}^{1 \times 2M-1}\rightarrow\mathbb{R}^{1 \times 2M-1}$ such that each training sample is mapped to its embedding as $f({\textbf{r}_{UL}}^{i})={{\hat{\textbf{r}}}_{DL}}^{i}$. Let ${\textbf{r}_{UL}}^{test}$ be the concatenated vector of an arbitrary UL CCM test point and $B_{\delta}({\textbf{r}_{UL}}^{test})$ be an open ball of radius $\delta$ around it
  \begin{equation}
  B_{\delta}({\textbf{r}_{UL}}^{test}) := \left\{\textbf{r}_{UL}\in \mathbb{R}^{2M-1}:\left\|{\textbf{r}_{UL}}^{test}-\textbf{r}_{UL}\right\|<\delta \right\}.
  \end{equation}
   Let $A^{UL}$ be the set of training samples within a $\delta$-neighborhood of ${\textbf{r}_{UL}}^{test}$ in $\mathbb{R}^{1 \times 2M-1}$
  \begin{equation} \label{A_UL}
  A^{UL} := \left\{{\textbf{r}_{UL}}^{i}:{\textbf{r}_{UL}}^{i} \in  B_{\delta}\left({\textbf{r}_{UL}}^{test}\right) \right\}.
  \end{equation}
   Denoting the support of the probability measure $\upsilon$ as $\mathcal{M} \subset \mathbb{R}^{1 \times 2M-1} $ , we define
  \begin{equation}
  \eta_{\delta} := \inf_{{\textbf{r}_{UL}}^{test} \in \mathcal{M}}\upsilon\left(B_{\delta}\left({\textbf{r}_{UL}}^{test}\right)\right)
  \end{equation} which is a lower bound on the measure of the open ball $B_{\delta}\left({\textbf{r}_{UL}}^{test}\right)$ around any test point.

\subsection{Theoretical Analysis Motivating the Proposed Method}

We now present a theoretical analysis of the regression problem of UL-to-DL CCM conversion via a mapping function $f(\cdot)$. We consider a setting with the following assumptions:
\begin{enumerate}
  \item  The function $f:\mathbb{R}^{1 \times 2M-1}\rightarrow\mathbb{R}^{1 \times 2M-1}$ is Lipschitz continuous with constant $L$; i.e., for any $\textbf{r}_1, \textbf{r}_2 \in \mathbb{R}^{1 \times 2M-1}$, we have $ \|f(\textbf{r}_1)-f(\textbf{r}_2)\| \leq L\| \textbf{r}_1-\textbf{r}_2 \| $. 
  \item The probability measure $\upsilon$ has a bounded support $\mathcal{M} \subset \mathbb{R}^{1 \times 2M-1}$. 
  \item For any $\delta>0$, the probability measure lower bound $\eta_{\delta}$ is strictly positive, i.e., $\eta_{\delta}>0$.
\end{enumerate}

We study the relation between the local geometries of the UL CCM and the DL CCM spaces in the following theorem.

\begin{theorem} \label{lemma-2}
Let ${\textbf{p}_{UL}}^{i} \in \mathbb{R}^{1 \times 2M-1}$ and ${\textbf{p}_{UL}}^{j} \in \mathbb{R}^{1 \times 2M-1}$ be obtained by concatenating the real and the imaginary parts of the first rows of two arbitrary UL CCMs. Assume that ${\textbf{p}_{UL}}^{i}$ and ${\textbf{p}_{UL}}^{j}$ are drawn i.i.d. from the probability measure $\upsilon$. Let ${\textbf{p}_{DL}}^{i}$ and ${\textbf{p}_{DL}}^{j}$ denote their DL counterparts.
If $\left\|{\textbf{p}_{UL}}^{i}-{\textbf{p}_{UL}}^{j} \right\| \leq 2\delta $, then, there exists a constant $K>0$ such that $\left\|{\textbf{p}_{DL}}^{i}-{\textbf{p}_{DL}}^{j}\right\| \leq K \left\|{\textbf{p}_{UL}}^{i}-{\textbf{p}_{UL}}^{j} \right\| \leq 2K\delta $, under the following assumptions:
 \begin{itemize}
\item The PAS, $p(\phi)$, is uniform.
\item $\delta$ is sufficiently small such that any two points $i$ and $j$ within the $\delta$-ball of a test point have very close mean angle of arrival (AoA) values, i.e., $\bar{v_i}-\bar{v_j}\approx0$.
\item The spread $\Delta$ of the AoA is constant and the same at each data point.
\end{itemize}
\end{theorem}

The proof of Theorem \ref{lemma-2} is given in \ref{Appendix_theorem_1_proof}.

\begin{remark}
\normalfont
Theorem \ref{lemma-2} suggests that, for the special case where the PAS is uniform and the angular spread of each user in a dataset is the same, if two points are close to each other in the UL CCM space, then they should be close to each other in the DL CCM space as well. In practice, the constant $K$ takes values close to $f_R$ in realistic settings. We demonstrate this with a numerical analysis in \ref{Appendix_K_constant}. Overall, Theorem \ref{lemma-2} provides useful insight for settings where a mapping function is to be learned between the spaces of UL CCMs and DL CCMs.
\end{remark}

For a sufficiently large dataset, i.e., for a sufficiently high $N$ value, the distance between a point in the dataset and its nearest neighbors shrinks considerably, so that the ball radius parameter $\delta$ becomes a small constant. In Theorem \ref{thm:some-theorem}, we consider such a setting and provide
an upper bound on the test error of the estimate of an arbitrary test point obtained via the interpolation function $f(\cdot)$.  

\begin{theorem}\label{thm:some-theorem}
Let the training sample set contain at least $N$ training samples $\{{\textbf{r}_{UL}}^{i}\}_{i=1}^{N}$ with ${\textbf{r}_{UL}}^{i} \sim \upsilon$. Let ${{\textbf{r}}_{UL}}^{test}$ be a test sample drawn from $\upsilon$ independently of the training samples.
Assume that the interpolation function $f:\mathbb{R}^{1 \times 2M-1}\rightarrow\mathbb{R}^{1 \times 2M-1}$ is a Lipschitz continuous function with Lipschitz constant $L$. Let $\epsilon>0$, $\frac{1}{N\eta_{\delta}} \leq a < 1$, and $ \delta > 0$  be arbitrary constants. Then, for a dataset with users  having uniform PAS ($p(\phi)$) with the same AS ($\Delta$), for sufficiently large $N$, with probability at least
 \begin{multline}   
  \label{prob_exp}
 \left(1-\exp\left(-2N(\left(1-a\right)\eta_{\delta})^2\right)\right) 
\left(1-2\sqrt{2M-1}\exp\left(-\frac{aN\eta_{\delta}\epsilon^2}{2L^2\delta^2}\right)\right)
 \end{multline} 
 the following inequality holds
 \begin{multline}  \label{the_upper_bound}
\left \|{{{{\textbf{r}_{DL}}^{test}}}-{f({\textbf{r}_{UL}}^{test})}}\right\| \\
\leq \frac{1}{|A^{UL}|} \sum_{i : {{\textbf{r}_{UL}}^{i}} \in A^{UL}}
 \left\| {\textbf{r}_{DL}}^{i} - f({\textbf{r}_{UL}}^{i}) \right\|
 + \left(L+K\right)\delta+\sqrt{2M-1}\epsilon.
 \end{multline} 
 \end{theorem}

\begin{figure}[!t]
\centering
\includegraphics[width=5.5in]{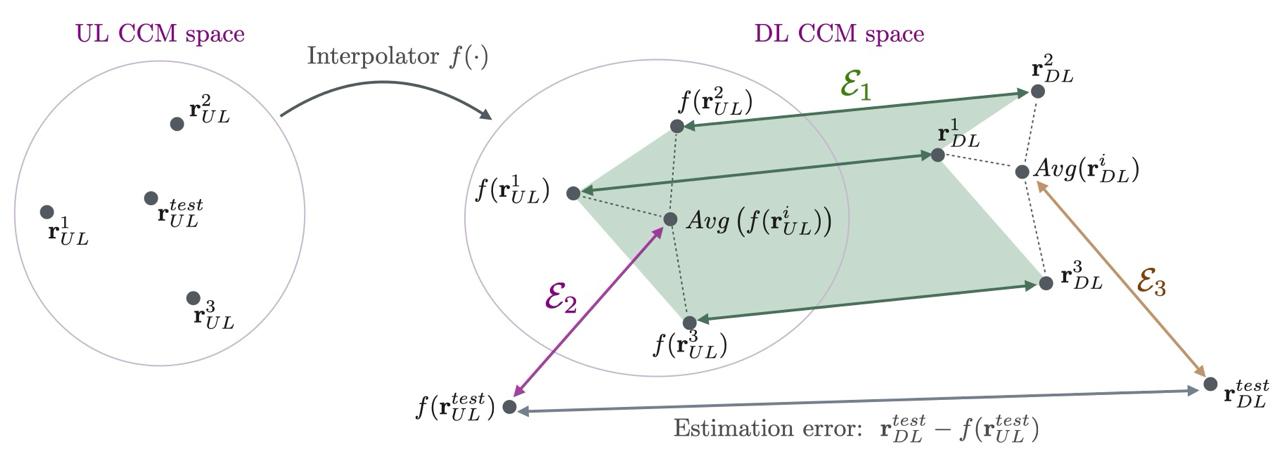}
\caption{Illustration of the proof of Theorem \ref{thm:some-theorem}. The estimation error  ${{{{\textbf{r}_{DL}}^{test}}}-{f({\textbf{r}_{UL}}^{test})}}$ of the interpolator $f(\cdot)$ is decomposed into the error components $\mathcal{E}_1$, $\mathcal{E}_2$, and $\mathcal{E}_3$. }
\label{fig_illus_thm2_proof}
\end{figure}
 
 The proof of Theorem \ref{thm:some-theorem} is given in \ref{Appendix_theorem1}. Our proof technique is based on decomposing the estimation error ${{{{\textbf{r}_{DL}}^{test}}}-{f({\textbf{r}_{UL}}^{test})}}$ into three components as illustrated in Figure \ref{fig_illus_thm2_proof}. The first error component (illustrated as $\mathcal{E}_1$ in Figure \ref{fig_illus_thm2_proof}) reflects how well the interpolator fits the training data within the local neighborhood of the interpolated point ${\textbf{r}_{UL}}^{test}$. The second error component ($\mathcal{E}_2$) is the deviation between the interpolator output $f({\textbf{r}_{UL}}^{test})$ and the average value of $ f({\textbf{r}_{UL}}^{i})$ for the training samples ${\textbf{r}_{UL}}^{i}$ in the local neighborhood of ${\textbf{r}_{UL}}^{test}$. This component is directly influenced by the Lipschitz constant $L$ of the interpolator, since the output of the interpolator exhibits smaller variation within a local region as its Lipschitz regularity improves. Finally, the third error component ($\mathcal{E}_3$) is the discrepancy between ${{\textbf{r}}_{DL}}^{test}$ and the value of $ {\textbf{r}_{DL}}^{i} $ averaged over the same local neighborhood, which is bounded through the local geometry preservation constant $K$ between the UL and the DL domains.   \\
 \begin{remark}
 \normalfont

Fixing the probability parameters $\delta>0$ and $\epsilon>0$ to sufficiently small constants,
one can see that the probability expression given in \eqref{prob_exp} approaches 1 at an exponential rate, as $N \rightarrow \infty$. Thus, it can be concluded that as $N \rightarrow \infty$, with probability approaching 1, the difference between the estimation error of a test point and the average estimation error of the training points within its ${\delta}$-neighborhood can be made as small as desired. (One can choose the $\delta$ parameter arbitrarily close to $0$, as $N \rightarrow \infty$.) From this result, one can conclude the following: 

 \begin{itemize}
  \item The smaller the average estimation error of the training points in the ${\delta}$-neighborhood of the test point is made by the algorithm that learns the function $f(\cdot)$, the smaller the upper bound on the estimation error of the test point gets. This can be achieved by arranging the objective function of the algorithm accordingly. 
  \item As the upper bound on the right hand side of \eqref{the_upper_bound} has a linear dependence on the Lipschitz constant $L$, our result suggests that the estimation error increases linearly with $L$. Meanwhile, the Lipschitz constant $L$ is not the only factor affecting the estimation error. In particular, the first term in the upper bound suggests that the interpolator should fit sufficiently well to the training samples $\{({\textbf{r}_{UL}}^{i},  {\textbf{r}_{DL}}^{i} ) \}$ in the local neighborhood of the interpolated point  ${\textbf{r}_{UL}}^{test}$. Also, the constant $K$ points to the importance of preserving the local geometry between the UL and DL spaces.  Equivalently, our result can be interpreted regarding the following trade-off between the Lipschitz constant $L$ and the training error: While one may reduce the training error to arbitrarily small values by increasing the complexity of $f(\cdot)$, this may come at the cost of learning a too irregular function with high Lipschitz constant $L$, resulting in poor generalization to new test data. Given all these factors, we conclude that an ideal interpolator $f(\cdot)$ must have a Lipschitz constant as small as possible, while also ensuring high data fitting accuracy and geometry preservation qualities. This key observation is behind the formulation of our objective function in this paper.
 \end{itemize}
 \end{remark}

 \section{DL CCM Estimation} \label{DL_CCM_Estimation_Section}
In this section, we propose a representation learning algorithm motivated by the theoretical analysis in the previous section for the problem of DL CCM estimation from UL CCM. 

\subsection{Problem Formulation} \label{Problem_Form_Subsection}
Let $\textbf{X} = [\left({\textbf{r}_{UL}}^{1}\right)^T ... \left({\textbf{r}_{UL}}^{N}\right)^T ]^T \in \mathbb{R}^{N \times (2M-1)} $ be the input training data matrix.
Let $\hat{\textbf{R}} = [\left({\hat{\textbf{r}}_{DL}}^{1}\right)^T ... \left({\hat{\textbf{r}}_{DL}}^{N}\right)^T]^T  \in \mathbb{R}^{N \times (2M-1)} $ be the embedding matrix, where ${\hat{\textbf{r}}_{DL}}^{i}=f({\textbf{r}_{UL}}^{i})$. 
Let $\textbf{R} = [\left({\textbf{r}_{DL}}^{1}\right)^T ... \left({\textbf{r}_{DL}}^{N}\right)^T ]^T \in \mathbb{R}^{N \times (2M-1)} $ be the output training data matrix, which is the DL counterpart of $\textbf{X}$.

Our aim is to find a function $f(\cdot)$ that approximates the training data sufficiently well, i.e., $f({\textbf{r}_{UL}}^{i}) = {\hat{\textbf{r}}_{DL}}^{i} \approx {\textbf{r}_{DL}}^{i} $, and preserves the nearest neighbors of each input vector in the embedding space, while mapping previously unseen UL CCMs (test data) to DL CCMs with low error. The interpolation problem can be formulated considering the following objectives. 

\subsubsection{Lipschitz regularity of the interpolation function}\label{lipschitz_subsubsection}
The interpolation function is of the form 
\begin{equation} \label{RBF_general_form}
    f({\textbf{r}_{UL}})=[f^{(1)}({\textbf{r}_{UL}}) \ f^{(2)}({\textbf{r}_{UL}}) \ ... \ f^{(2M-1)}({\textbf{r}_{UL}})].
\end{equation}

\noindent
In our work, we choose $f(\cdot)$  as a Gaussian radial basis function (RBF) interpolator due to its well-studied properties \cite{piret2007analytical}:
The smoothness of RBF interpolators is one of the main factors affecting its accuracy \cite{piret2007analytical}, where the choice of Gaussian kernels ensures infinitely continuously differentiable mappings and leads to a super-spectrally accurate interpolator \cite{piret2007analytical}.
Another advantageous property of RBF interpolators is that their Lipschitz regularity can be analytically studied. In \cite{ornek2019nonlinear}, the Lipschitz constant of Gaussian RBF interpolators is provided in closed-form, which can be used to control the regularity of the interpolation function to be learned. While the Lipschitz constants of other interpolators such as cubic and polyharmonic splines are also analytically tractable, Gaussian RBFs have the additional advantage of incorporating a tunable scale parameter $\sigma_{rbf}$ in their formulation. By directly controlling the rate of decay of Gaussian kernels, the scale parameter $\sigma_{rbf}$ provides an extra important degree of freedom when seeking a balance between the fidelity of the embedding to the available training data and the Lipschitz regularity of the interpolator.
For the extension of the embedding, we thus consider a mapping $f({\textbf{r}_{UL}})$ whose  $k^{th}$ element is of the form
\begin{equation} \label{RBF_each_element}
    f^{(k)}({\textbf{r}_{UL}}) = \sum_{i=1}^{N} C_{ik} \ e^{-\frac{\|{\textbf{r}_{UL}}-{\textbf{r}_{UL}}^{i}\|^2}{\sigma_{rbf}^2} }
\end{equation}
for $k\in \{1,...,2M-1\}$. Here $C_{ik}$ are the interpolator coefficients and $\sigma_{rbf}$ is the scale parameter of the Gaussian RBF kernel.

A Lipschitz constant for the Gaussian RBF interpolation function is provided in \cite{ornek2019nonlinear} as
\begin{equation} \label{Lipschitz_Constant_RBF}
    L = \sqrt{2}e^{-1/2}\sqrt{N}\sigma_{rbf}^{-1}\|\textbf{C}\|_F ,
\end{equation}
where $\textbf{C} \in \mathbb{R}^{N \times (2M-1)}$ is the matrix containing the interpolator coefficient $C_{ik}$ in its $(i,k)^{th}$ element, $i \in \{ 1, ..., N\}$, $k \in \{ 1, ..., 2M-1\}$. The matrix $\textbf{C}$ is obtained as 
\begin{equation} \label{interpolator_coeffs_C_mtx}
    \textbf{C}=\boldsymbol \Psi^{-1} \hat{\textbf{R}} 
\end{equation}
by learning a mapping $\hat{\textbf{R}}$ from the training data matrix  $\textbf{X}$, where $\boldsymbol \Psi \in \mathbb{R}^{N \times N}$ is the RBF kernel matrix, whose $(i,j)^{th}$ element is $e^{-\frac{\|{\textbf{r}_{UL}}^{i}-{\textbf{r}_{UL}}^{j}\|^2}{\sigma_{rbf}^2}}$.

From Theorem \ref{thm:some-theorem}, the Lipschitz constant $L$ of the interpolator $f(\cdot)$ should be small so as to reduce the error upper bound in \eqref{the_upper_bound}, which improves the generalization of the embedding to test data. Considering this along with \eqref{Lipschitz_Constant_RBF}, we propose to minimize the following terms when learning the embedding coordinates and the function parameters of the RBF interpolator:
\begin{itemize}
    \item $\sigma_{rbf}^{-2} $
    \item $\| \textbf{C}\|_{F}^{2}=\| \boldsymbol \Psi^{-1} \hat{\textbf{R}}\|_{F}^{2}=tr(\hat{\textbf{R}}^{T}\boldsymbol \Psi^{-2}\hat{\textbf{R}})$
\end{itemize}

\subsubsection{Preservation of the local geometry between the UL/DL CCM spaces}\label{local_geometry_subsubsection}
Due to the angular reciprocity, there is an inherent similarity between the UL CCM and the DL CCM of the same user, despite the lack of an explicit function relating them. On the other hand, Theorem \ref{lemma-2} indicates that the neighboring points in the UL space are positioned closer with increasing amounts of training data, in which case the distance between their DL counterparts can also be bounded proportionally.
We aim to preserve the local geometry of UL CCMs in the embedding space. In order to  achieve this, we minimize the term
\begin{equation} \label{Laplacian_eigenmap}
    \sum_{i,j=1}^{N} (\textbf{W})_{ij} \| \hat{\textbf{r}}_{DL}^{i}-\hat{\textbf{r}}_{DL}^{j} \|^{2}=tr(\hat{\textbf{R}}^{T}\textbf{L}\hat{\textbf{R}}),
\end{equation}
where $\textbf{W}$ is a weight matrix whose $(i,j)^{th}$ entry is given by $(\textbf{W})_{ij}=e^{-\frac{\|{\textbf{r}_{UL}}^{i}-{\textbf{r}_{UL}}^{j}\|^2}{\theta^2}}$ (for a scale parameter $\theta$), $\textbf{L}=\textbf{D}-\textbf{W}$ is the Laplacian matrix, and $\textbf{D}$ is the diagonal degree matrix with $i^{th}$ diagonal entry $(\textbf{D})_{ii}=\sum_{j} (\textbf{W})_{ij}$. The weights in the weight matrix are selected according to the pairwise distances between data pairs, i.e., $\| {\textbf{r}_{UL}}^{i} - {\textbf{r}_{UL}}^{j} \|$ for $i,j \in \{1,...,N\}, i \neq j$.
In this way, for nearby $({\textbf{r}_{UL}}^{i}, {\textbf{r}_{UL}}^{j})$ pairs with strong edge weights, a high penalty is applied to
the action of mapping $\hat{\textbf{r}}_{DL}^{i}$ and $\hat{\textbf{r}}_{DL}^{j}$ far from one another, which preserves the structure of the local neighborhoods between the UL and the DL domains \cite{belkin2003laplacian}. The equality in \eqref{Laplacian_eigenmap} is shown in \cite{belkin2003laplacian}.\\

\subsubsection{UL/DL CCM pairs in the training dataset}\label{train_datapairs_subsubsection}
As we aim to learn a function that maps UL CCMs to their corresponding DL CCMs, the UL-DL CCM pairs in the training dataset are also incorporated into our optimization problem. Instead of employing hard data fidelity constraints, in order to achieve better noise tolerance we prefer the quadratic penalty term given by \[ \| \hat{\textbf{R}}-\textbf{R} \|_{F}^{2}. \] 

\subsubsection{Overall problem} \label{overall_problem_subsubsection}
We finally combine the above terms to form our overall objective function as
\begin{equation} \label{Overall_Problem}
    \min_{\hat{\textbf{R}},\sigma_{rbf}} tr(\hat{\textbf{R}}^{T}\textbf{L}\hat{\textbf{R}})
    +\mu_{1}tr(\hat{\textbf{R}}^{T}\boldsymbol \Psi^{-2}\hat{\textbf{R}}) 
    +\mu_{2}\sigma_{rbf}^{-2}+\mu_{3}\| \hat{\textbf{R}}-\textbf{R} \|_{F}^{2},
\end{equation}
where $\mu_{1}$, $\mu_{2}$ and $\mu_{3} $ are positive weights to determine the relative importance of each term in the objective function. In particular, the regularization term $tr(\hat{\textbf{R}}^{T}\boldsymbol \Psi^{-2}\hat{\textbf{R}})$ is equal to the energy $\| \mathbf{C}\|_F^2$ of the interpolator coefficients, which tends to increase when $\sigma_{rbf}$ is too high. On the other hand, the regularization term $\sigma_{rbf}^{-2}$ prevents the scale parameter to take too small values, which ensures the expressive power of the interpolator. These two regularization terms, together with the data fidelity and geometry preservation terms, provide a basis for learning an efficient model that finds a suitable balance between the smoothness and the expressiveness of the interpolator.\\

\subsection{Solution of the Problem}
The optimization problem defined above is not jointly convex in $\hat{\textbf{R}}$ and $\sigma_{rbf}$. We employ an alternating optimization method, where one of the parameters is fixed while the other one is optimized in an alternative fashion at each iteration. This alternation is continued until convergence or the maximum number of iterations is reached. 
The objective function is nonconvex. Therefore, it is difficult to provide a theoretical guarantee for the convergence of the solution. Nevertheless, since
$\mu_{1}$, $\mu_{2}$ and $\mu_{3} $ are positive numbers, the objective function in \eqref{Overall_Problem} always converges to a nonnegative value.

\textbf{Optimization of $\hat{\textbf{R}}$:} When $\sigma_{rbf}$ is fixed, the optimization problem in \eqref{Overall_Problem} becomes 
\begin{equation} \label{optimization_of_Y}
    \min_{\hat{\textbf{R}}} tr(\hat{\textbf{R}}^{T}\textbf{L}\hat{\textbf{R}})
    +\mu_{1}tr(\hat{\textbf{R}}^{T}\boldsymbol \Psi^{-2}\hat{\textbf{R}}) 
    +\mu_{3}\| \hat{\textbf{R}}-\textbf{R} \|_{F}^{2}
\end{equation}
where the objective function is quadratic and convex. The closed form solution of the problem in \eqref{optimization_of_Y} is given by 
\begin{equation} \label{Solution_of_Y}  \hat{\textbf{R}}^*=\mu_{3}\left(\textbf{A}+\mu_{3}\textbf{I}\right)^{-1}\textbf{R},
\end{equation}
where $\textbf{A}=\textbf{L}+\mu_{1}\boldsymbol \Psi^{-2}$. The eigenvalues of a graph Laplacian matrix are always nonnegative, i.e., the Laplacian matrix is a positive semidefinite matrix.
Therefore, the matrix $\left(\textbf{A}+\mu_{3}\textbf{I}\right)$ is always invertible.

\textbf{Optimization of $\sigma_{rbf}$:} When $\hat{\textbf{R}}$ is fixed, the optimization problem in \eqref{Overall_Problem} can be rewritten as
\begin{equation} \label{optimization_of_sigma}
    \min_{\sigma_{rbf}}\mu_{1}tr(\hat{\textbf{R}}^{T}\boldsymbol \Psi^{-2}\hat{\textbf{R}})+\mu_{2}\sigma_{rbf}^{-2}.
\end{equation}
Although nonconvex, this problem involves the optimization of a single scalar variable $\sigma_{rbf}$, which can be solved via an exhaustive search of $\sigma_{rbf}$ in a reasonable interval. 

Our solution algorithm is summarized in Algorithm 1.
\begin{algorithm}[htbp]
\SetKwInOut{Input}{input}\SetKwInOut{Output}{output}
\SetAlgoLined
\Input{Training data matrices $\textbf{X}$ and $\textbf{R}$ }
 \textbf{Initialization:} \\
 Construct the graph Laplacian matrix $\textbf{L}$ and the RBF kernel matrix $\boldsymbol \Psi$ \\
 Assign weight parameters $\mu_{1}$, $\mu_{2}$ and $\mu_{3}$ and initial values of $\sigma_{rbf}$ and $\hat{\textbf{R}}$
 
 \Repeat{convergence of the objective function or the maximum iteration number is reached}{
Fix $\sigma_{rbf}$ and optimize $\hat{\textbf{R}}$ via \eqref{Solution_of_Y}

Fix $\hat{\textbf{R}}$ and optimize $\sigma_{rbf}$ via \eqref{optimization_of_sigma}
 }
 \Output{Kernel scale parameter $\sigma_{rbf}$, embedding matrix $\hat{\textbf{R}}$  }
 \caption{DL CCM Interpolation via Gaussian RBF Kernel}
\end{algorithm}

After learning the embedding matrix $\hat{\textbf{R}}$ and the kernel scale parameter $\sigma_{rbf}$ with Algorithm 1, one can calculate the interpolator coefficient matrix $\textbf{C}$ from \eqref{interpolator_coeffs_C_mtx}. Thus, using \eqref{RBF_general_form} and \eqref{RBF_each_element}, one can estimate the DL CCM of a new test sample that is not in the training dataset by using its UL CCM.
\\

The integral of the PAS over all angles is known to be $1$; however,
we do not enforce such a normalization when learning the embedding and the kernel scale parameter. 
For this reason, once we obtain the estimate $\hat{\textbf{r}}_{DL}$, we normalize it by setting its first entry  to $\hat{\textbf{r}}_{DL}(1)=1$.

\subsection{Complexity Analysis}
The main factors that determine the complexity of our algorithm are the optimization problems given in \eqref{optimization_of_Y} and \eqref{optimization_of_sigma}, which are solved in an alternating fashion.
The complexity of constructing the matrices $\textbf{L}$ and $\boldsymbol \Psi$ is $\mathcal{O}(MN^2)$, where $N$ is the number of training samples in the dataset. The matrix inversion operations in \eqref{Solution_of_Y} and \eqref{optimization_of_sigma} are of complexity $\mathcal{O}(N^3)$, which is the decisive part of the complexity analysis in a typical scenario where $M<N$. Hence, the overall complexity of our algorithm is $\mathcal{O}(N^3)$.

Once the training is completed, the Gaussian RBF interpolation function can be directly used to find the DL CCMs of new data. The complexity of finding an estimate of the DL CCM using our function is $\mathcal{O}(M^2N)$, since each element of the embedding vector of size $\left(2M-1\right)$ involves the processing of $\left(2M-1\right)$-dimensional vectors at $N$ center locations. \cite{roussos2005rapid}.

\subsection{Discussion about the Generalizability of the Algorithm to Different Scenarios}
In this study, we have proposed a theoretical analysis and an algorithm by considering a ULA antenna setting and the Toeplitz CCM structure arising from this topology. In this section, we briefly discuss the potential extension of our method to other scenarios. 
As an example, we consider the uniform rectangular array (URA) BS antennas, which is another commonly employed setting. In this case, the CCM is block Toeplitz with Toeplitz blocks (BTTB) \cite{10189874}. This means that the CCM is block Hermitian and block Toeplitz, where the diagonal block consists of the repetition of a Hermitian Toeplitz matrix and the off-diagonal blocks are made up of matrices that are only Toeplitz. Hence, the matrix tiled along the diagonal block can be described by its first row, while the other matrices can be characterized by their first rows and first columns. If the URA has $M_r M_c = M$ antennas, the overall CCM can be described by $(2M_r-1)(2M_c-1)=4M-2(M_r+M_c)+1$ elements. 
In particular, it is typical for structured antenna array geometries (such as ULA and  URA) to result in CCMs that can be described by a relatively small number of model parameters. Note that our objective function in \eqref{Overall_Problem} can be adapted to these settings in a straightforward way, where the number of parameters is the only change in the problem.
Our theoretical analysis relies on ULA-specific properties. Extending it to the URA case is nontrivial and requires future investigation.

Regarding the PAS, we have theoretically examined only the uniform PAS scenario; nevertheless, included also other PAS shapes such as Laplacian and Gaussian distributions in our simulations. The current scope of our study is limited to a simplified propagation scenario, as opposed to more complex multi-cluster massive MIMO scenarios as in \cite{kurt2023robust}. In \cite{kurt2023robust}, the overall CCM is modeled as the sum of the CCMs of each cluster. All clusters have uniform PAS similarly to our model. This setting differs from ours in the sense that summing up all these uniform PASs over different angular ranges, one would obtain a different, probably a random, PAS shape. One may potentially use specific properties of this sum of uniform-PAS-CCMs structure  for adapting our algorithm to communication scenarios with multi-cluster CCMs. Regarding these points, the investigation of the performance of our method in multi-cluster scenarios remains as a promising future extension of our study.

Another frequently encountered practical scenario is when the training data set contains noisy uplink CCMs. We note that, although our method does not explicitly perform denoising, it is able to implicitly handle data noise to some extent:  The term $\| \hat{\textbf{R}}-\textbf{R} \|_{F}^{2}$ in our objective function \eqref{Overall_Problem} represents the fidelity of the computed embedding $\hat{\textbf{R}}$ to the training data $\textbf{R}$. In a setting where the training data set  $\textbf{R}$  is noisy, our algorithm has the flexibility to compute an embedding $\hat{\textbf{R}}$ that may deviate from the noisy coordinates as required, by suitably adjusting the weight parameters $\mu_1$, $\mu_2$ and $\mu_3$ to set the relative importance of the data fidelity term. In scenarios with severe noise, one may also couple our method with a preliminary denoising module for achieving robustness.

\section{Simulations} \label{Simulation_Section}
In this section, we evaluate the performance of our algorithm with simulations, based on the simulation setup reported in Table \ref{sim_param_table}.  We first observe the behavior of the objective function and that of the estimation performance of our method throughout the iterations. 
Next, we conduct tests to study how the performance of our method varies with algorithm hyperparameters.
Finally, we compare the performance of our method to that of some baseline methods in the literature.

\begin{table}[!h]
\caption{Simulation Parameters}
\label{sim_param_table}
\centering
\begin{tabular}{ |p{3.7cm}|p{3.5cm}|  }
 \hline
 Carrier Frequencies& $f_{UL}=1.95$ GHz, $f_{DL} \ \ \ = \ \ \ 2.14$ \ GHz \\
 \hline
 Number of Base Station Antennas (M)   & One of the following: $\{32,64,128,256\}$ \\
 \hline
 Dataset Size (Training and Test)&   $500$ \\
 \hline
 Training/Test Data Ratio & $80 \% / 20 \%$  \\
 \hline
 $\mu_1$, $\mu_2$, $\mu_{3}$    &  $0.1,3\times10^5,100$\\
 \hline
 SNR&   20 dB  \\
 \hline
\end{tabular}
\end{table}

Users are considered to have uniform PAS (unless stated otherwise) with mean AoAs uniformly distributed in $[-\pi, \pi]$. The spread of AoAs of users are drawn from $[5\degree,15\degree]$ uniformly. The carrier frequencies of uplink and downlink channels in Table \ref{sim_param_table} are chosen according to \cite{3gppTS36101}.

Let us denote the true value of a DL CCM by $\textbf{R}_{DL}$ and its estimate by $\hat{\textbf{R}}_{DL}$. The following three error metrics are used to compare the performance of the proposed algorithm with benchmark methods:
\begin{enumerate}
  \item Normalized Mean Square Error (NMSE): NMSE is used to measure the average error in each entry of a CCM, which is defined as
  \begin{equation} \label{NMSE}
     \text{NMSE} = \mathop{\mathbb{E}} \Biggl\{\frac{\|\textbf{R}_{DL}-\hat{\textbf{R}}_{DL}\|_{F}^{2}}{\|\textbf{R}_{DL}\|_{F}^{2}}\Biggr\}. 
  \end{equation}
  \item Correlation Matrix Distance (CMD): This metric defined in \cite{herdin2004mimo} is used to quantify the deviation between the direction of the true DL CCM and that of its estimate. The CMD is given by
  \begin{equation}
       \text{CMD} = \mathop{\mathbb{E}} \Biggl\{1-\frac{tr(\textbf{R}_{DL} \hat{\textbf{R}}_{DL})}{\|\textbf{R}_{DL}\|_{F} \|\hat{\textbf{R}}_{DL}\|_{F}}\Biggr\}.
  \end{equation}
  \item Deviation Metric (DM):
  In \cite{10189874}, the following metric is used to measure the deviation in the principal eigenvector of the estimated DL CCM, which is useful in beamforming applications: 
  \begin{equation}
      \text{DM} = 1 - \frac{tr(\textbf{v}^H \textbf{R}_{DL} \textbf{v})}{\Gamma_{max}},
  \end{equation}
  where $\Gamma_{max}$ is the largest eigenvalue of $\textbf{R}_{DL}$ and $\textbf{v}$ is the eigenvector corresponding to the largest eigenvalue of $\hat{\textbf{R}}_{DL}$.\\
\end{enumerate}

\subsection{Simulation Setup}

The dataset is constructed similarly to the setting in \cite{10189874} as described below. The following steps are followed for all UL CCMs in the dataset and for the DL CCMs in the training dataset. DL CCMs in the test set are constructed via only Step 1 in order to obtain an ideal ground truth dataset for performance comparisons of our algorithm with the benchmark methods.

\begin{enumerate}
 \item CCMs are calculated using the formula in \eqref{CCM_Formula}.
  \item UL and DL channel realizations are constructed from the CCMs as
    \begin{equation} \label{channel_realizations}      \left(\textbf{h}_x^{k}\right)^{c}=\left(\textbf{R}_x^{k}\right)^{1/2}
        \left(\textbf{w}_{x}^{k}\right)^{c}, c=1,...,N_{ch}, 
        x \in \{UL,DL\}, 
    \end{equation}
    where $\left(\textbf{w}_{x}^{k}\right)^{c} \sim \mathcal{CN}(\textbf{0},\,\textbf{I})$, the matrix $\textbf{R}_x^{k}$ is the CCM of user $k$ (either UL or DL, specified by $x$) and $N_{ch}$ is the number of channel realizations. $N_{ch}$ is taken as $2M$ in the simulations unless it is specified otherwise.
    \item The noisy channel estimates obtained after the training phase with pilot signals are modeled and generated as 
    \begin{equation} \left(\hat{\textbf{h}}_{x}^{k}\right)^{c}=\left(\textbf{h}_{x}^{k}\right)^{c}+\left(\textbf{n}_{x}^{k}\right)^{c}, c=1,...,N_{ch},
    x \in \{UL,DL\} 
    \end{equation} 
    where $\left(\textbf{n}_{x}^{k}\right)^{c}\sim \mathcal{CN}(\textbf{0},\,\sigma_{noise}^{2}\textbf{I})$ and $\left(\hat{\textbf{h}}_{x}^{k}\right)^{c}$ is the noisy channel estimate of the $c^{th}$ channel realization.
   The signal-to-noise ratio (SNR) $tr(\textbf{R}_{UL}^{k}) / \sigma_{noise}^2$ for this pilot signaling setup is taken to be $20$ dB as in \cite{10189874}, unless it is explicitly said to be taken differently.
    \item The sample covariance for user $k$ is then given by
    \begin{equation} \label{sample_covariance}
       \hat{\textbf{R}}_{x}^{k} = \frac{1}{N_{ch}} \sum_{c=1}^{N_{ch}}  \left(\hat{\textbf{h}}_{x}^{k}\right)^{c} \left(\hat{\textbf{h}}_{x}^{k}\right)^{{c}^{H}} -\sigma_{noise}^2\textbf{I},
       x \in \{UL,DL\}. 
    \end{equation}
    \item Due to the ULA antenna structure at the BS and the WSSUS model, the CCMs are Toeplitz, Hermitian and PSD, which is used for the correction of the sample covariance found in \eqref{sample_covariance}.  The sample covariance matrices are projected onto the set of Toeplitz, Hermitian and PSD matrices with the alternative projection method proposed in \cite{grigoriadis1994application}.
    The projection method solves the optimization problem
    \begin{equation}
        \Tilde{\textbf{R}}_{x}^{k}=\arg \min_{\textbf{X} \in T_+^M} \|\textbf{X}-\hat{\textbf{R}}_{x}^{k} \|^2
    \end{equation}
    where $T_+^M$ is the set of $M \times M$ Toeplitz, Hermitian and PSD matrices.
    \item The matrices estimated in the previous step are normalized such that their $(1,1)^{th}$ element is 1. This is done due to the fact that the PAS of the CCMs are normalized to 1.

\end{enumerate}

\subsection{Stability and Sensitivity Analysis} \label{Sensitivity_Section}
\begin{figure}[!t]
\centering
\includegraphics[width=3in]{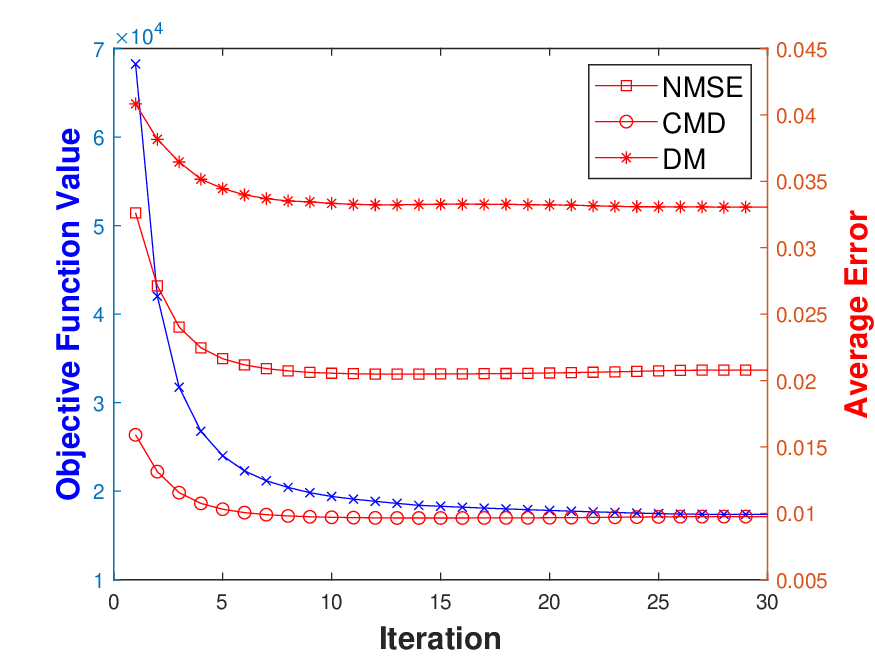}
\caption{The variation of the objective function and the average error throughout the iterations}
\label{stability}
\end{figure}

First, we study the change in the objective function and the change in the average NMSE of DL CCMs learned by our algorithm throughout the iterations. For $M=64$ base station antennas,
we repeat the experiments for 25 i.i.d. datasets.
The average objective function and error values are presented in Figure \ref{stability}. In Figure \ref{stability}, one can see that the objective function decreases throughout the iterations, which is expected because the algorithm updates both the embedding and the kernel scale parameter in such a way that the objective function never increases. The average NMSE, CMD and DM exhibit a similar decreasing trend consistent with the behavior of the objective function, which suggests that our proposed objective function captures the performance goal of our algorithm well.

Next, we conduct a sensitivity analysis in order to examine the effect of the hyperparameters ($\mu_1, \mu_2, \mu_3$) on the performance of our algorithm.
Tables \ref{tab:table_mu1_mu2} and \ref{tab:table_mu1_mu3} show the NMSE values of the DL CCM estimates of our algorithm for several ($\mu_1, \mu_2, \mu_3$) combinations. For each ($\mu_1, \mu_2, \mu_3$), we repeat the experiments for 10 i.i.d. datasets, where the base station has $M=64$ antennas. In the experiments of each table, the fixed parameter among $\mu_1, \mu_2, \mu_3$ is manually set to a suitable value and the other two parameters are swept within the indicated range.

Table \ref{tab:table_mu1_mu2} points to the necessity of the terms regulating the Lipschitz constant in the objective function, since the performance improves as $\mu_1$ and $\mu_2$ increase together from $0$ up to around $\mu_{1}=10^{-1}$ and $\mu_{2}=3 \times 10^{5}$. Beyond these values, the performance deteriorates as the terms related to the Lipschitz constant begin to dominate the objective function in \eqref{Overall_Problem}, which reduces the impact of the data fidelity terms. This causes the mappings of the training points to deviate from their true values and ultimately leads to performance degradation. 

Table \ref{tab:table_mu1_mu3} reports the performance for different weight combinations for the Lipschitz continuity of the interpolator and the data fidelity. The ratio between $\mu_1$ and $\mu_2$ is fixed to a suitable value chosen based on Table \ref{tab:table_mu1_mu2}. The results in Table \ref{tab:table_mu1_mu3} show that as $\mu_3$ gets smaller, the average NMSE increases drastically. However, it also shows that $\mu_1$ (and also $\mu_2$) should be chosen as positive numbers to improve the performance. The performance seems to be more sensitive to the data fidelity term than the Lipschitz continuity terms. 

\begin{table}[!t]
\caption{The Variation of the NMSE with the Hyperparameters 
$\mu_{1}$ and $\mu_{2}$ for Fixed $\mu_{3}=100$ \label{tab:table_mu1_mu2}}
\centering
\begin{adjustbox}{width=\columnwidth,center}
\begin{tabular}{|c|c|c|c|c|c|c|c|c|}
\hline
\backslashbox{$\mu_{2}$}{$\mu_{1}$} & $0$ & $10^{-4}$ & $10^{-3}$ & $10^{-2}$ & $10^{-1}$ & $1$& $10^{1}$ & $10^{2}$\\
\hline
$0$ & 0.0463 & 0.0390 & 0.0390 & 0.0389 & 0.0376 & 0.0345 & 0.0309 & 0.0402\\
\hline
$3 \times 10^{-1}$ & 0.0463 & 0.0348 & 0.0378 & 0.0387 & 0.0376 & 0.0345 & 0.0309 & 0.0402\\
\hline
$3 \times 10^{1}$ & 0.0463 & 0.0313 & 0.0320 & 0.0344 & 0.0361 & 0.0343 & 0.0309 & 0.0402\\
\hline
$3 \times 10^{3}$ & 0.0463 & 0.0349 & 0.0325 & 0.0307 & 0.0297 & 0.0298 & 0.0300 & 0.0403\\
\hline
$3 \times 10^{5}$ & 0.0463 & 0.0265 & 0.0221 & 0.0194 & 0.0201 & 0.0238 & 0.0319 & 0.0452\\
\hline
$3 \times 10^{7}$ & 0.0463 & 0.0265 & 0.0221 & 0.0194 & 0.0208 & 0.0313 & 0.0540 & 0.0796\\
\hline
$3 \times 10^{9}$ & 0.0463 & 0.0265 & 0.0221 & 0.0194 & 0.0208 & 0.0313 & 0.0540 & 0.1022\\
\hline
$3 \times 10^{11}$ & 0.0463 & 0.0265 & 0.0221 & 0.0194 & 0.0208 & 0.0313 & 0.0540 & 0.1022\\
\hline
\end{tabular}
\end{adjustbox}
\end{table}

\begin{table}[!t]
\caption{The Variation of the NMSE with the Hyperparameters 
$\mu_{1}$ and $\mu_{3}$ for $\mu_{2}=3\times10^{6}\mu_{1}$ \label{tab:table_mu1_mu3}}
\centering
\begin{adjustbox}{width=\columnwidth,center}
\begin{tabular}{|c|c|c|c|c|c|c|c|c|}
\hline
\backslashbox{$\mu_{3}$}{$\mu_{1}$} & $0$ & $10^{-4}$ & $10^{-3}$ & $10^{-2}$ & $10^{-1}$ & $1$& $10^{1}$ & $10^{2}$\\
\hline
$10^{-1}$ & 0.2093 & 0.2248 & 0.2448 & 0.2550 & 0.2739 & 0.3643 & 0.6225 & 0.8106\\
\hline
$1$ & 0.0722 & 0.0704 & 0.0682 & 0.0730 & 0.0858 & 0.1252 & 0.2766 & 0.6016\\
\hline
$10^{1}$ & 0.0357 & 0.0315 & 0.0277 & 0.0241 & 0.0347 & 0.0566 & 0.1043 & 0.2657\\
\hline
$10^{2}$ & 0.0463 & 0.0324 & 0.0325 & 0.0275 & 0.0201 & 0.0311 & 0.0540 & 0.1022\\
\hline
$10^{3}$ & 0.0496 & 0.0330 & 0.0331 & 0.0335 & 0.0283 & 0.0199 & 0.0308 & 0.0538\\
\hline
$10^{4}$ & 0.0500 & 0.0331 & 0.0331 & 0.0332 & 0.0336 & 0.0284 & 0.0198 & 0.0307\\
\hline
$10^{5}$ & 0.0500 & 0.0331 & 0.0331 & 0.0331 & 0.0332 & 0.0336 & 0.0284 & 0.0198\\
\hline
\end{tabular}
\end{adjustbox}
\end{table}

\begin{figure}[!t]
\centering
\includegraphics[width=3.5in]{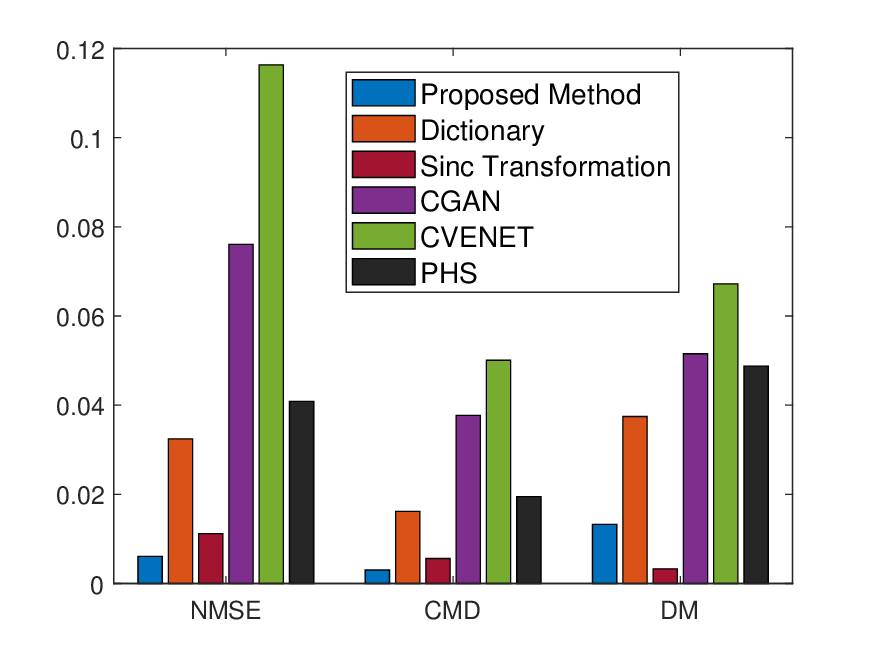}
\caption{Average NMSE, CMD and DM values of the compared methods  for a perfect dataset with CCMs of a system with $M=256$ base station antennas}
\label{error_bar_all_methods}
\end{figure}

\begin{figure}[!t]
\centering
\subfloat[]{\includegraphics[width=2.5in]{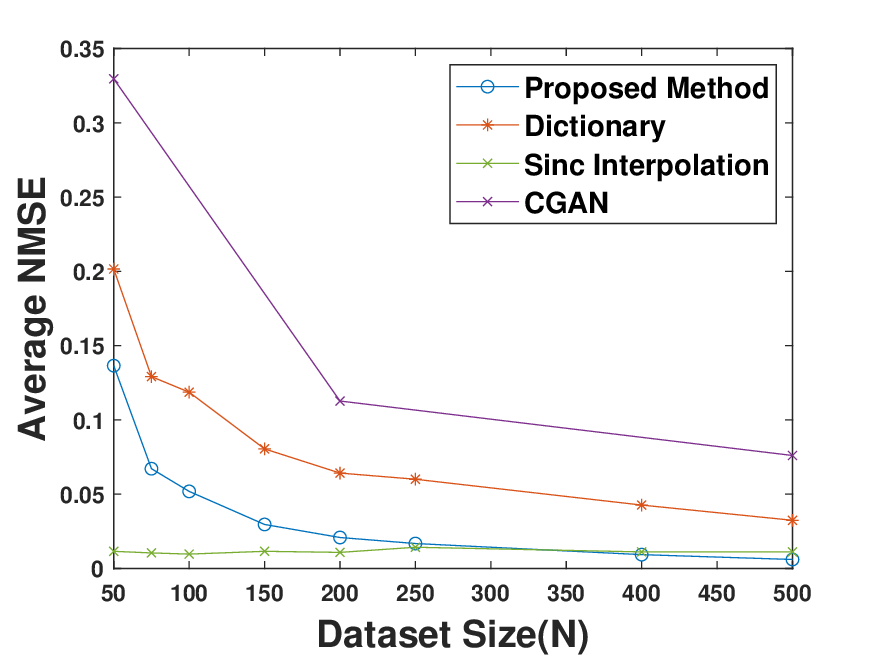}
\label{N_vs_NMSE}}
\hfil
\subfloat[]{\includegraphics[width=2.5in]{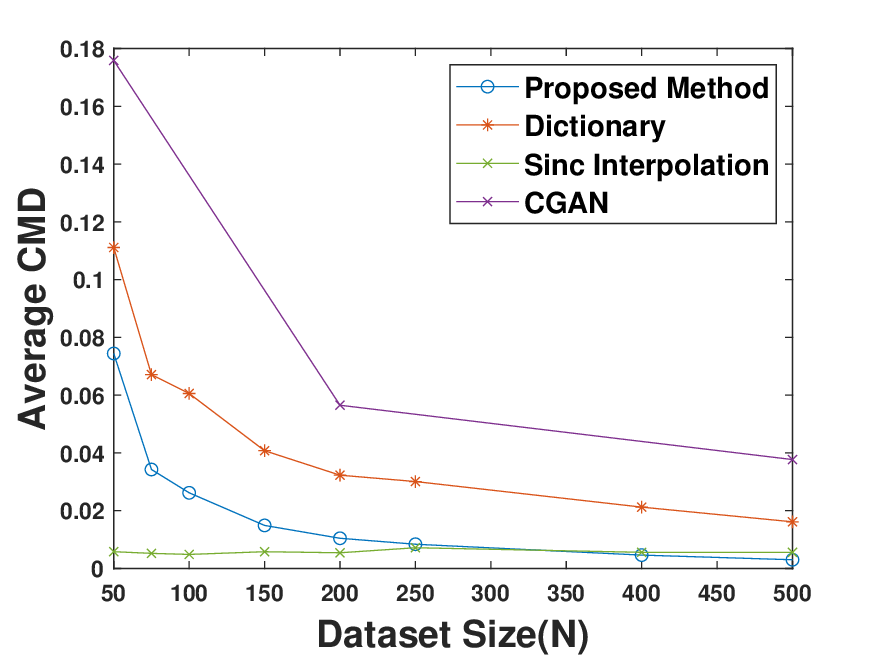}
\label{N_vs_CMD}}
\hfil
\subfloat[]{\includegraphics[width=2.5in]{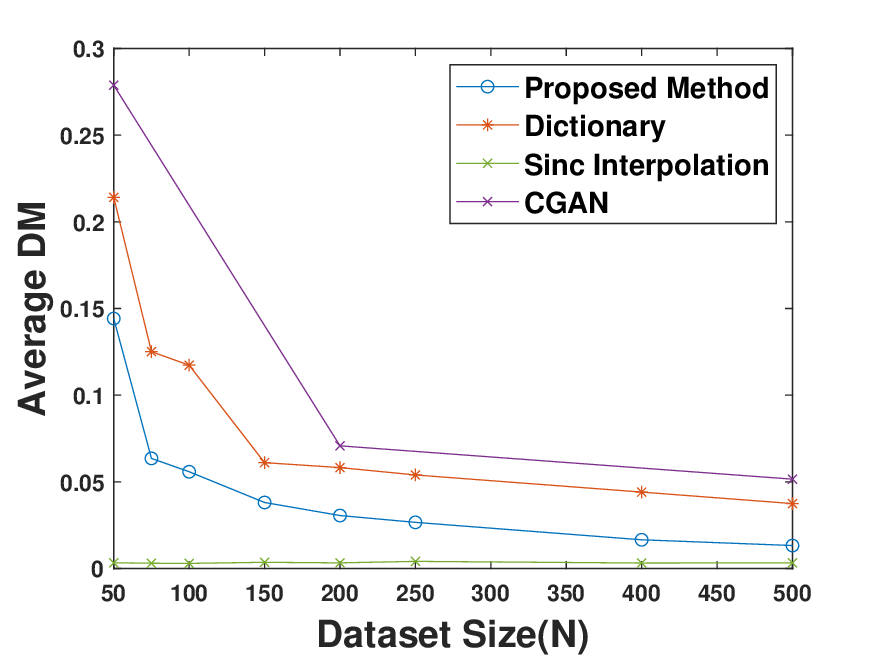}
\label{N_vs_DM}}
\caption{The variation of the errors of the compared methods with dataset size, (a) NMSE (b) CMD (c) DM}
\label{N_vs_Error}
\end{figure}

\subsection{Algorithm Performance}
In this section, we compare the average NMSE, CMD and DM values  of our method to those of the following five benchmark methods: (1) The polyharmonic splines method (PHS) described in \cite{FlyerFBB16}, (2) the dictionary method in \cite{decurninge2015channel}, (3) the sinc transformation method in \cite{10189874}, (4) the CGAN model in \cite{banerjee2022downlink}, (5) the variational autoencoder model CVENET \cite{linfu2021deep}.
The PHS method is a classical and well-known interpolation technique, which we find instructive to compare with our interpolation algorithm employing graph regularization via Gaussian RBF kernels. The dictionary method is a well-established algorithm in the literature, which bears similarity to our method in the sense that both methods rely on the idea of preserving the neighborhood relationships between the UL CCM and the DL CCM domains. The recently proposed sinc transformation method presents a simple  solution based on signal processing. The CGAN and CVENET methods are recent deep learning-based  approaches for   estimating CCMs, which is a strategy that has gained popularity for solving communications problems. We prefer to include these methods in our experiments due to their relevance and recency: Both methods address the UL-to-DL CCM conversion task using deep learning and represent CCMs as RGB images, making them suitable baselines for comparison with our approach. They both benefit from the encoder-decoder structure to learn the common features between the UL and DL CCMs, which originate from the same PAS. The CGAN method leverages the adversarial structure of the conditional generative adversarial network by simultaneously training a generator that outputs fake images and a discriminator that distinguishes the fake images from the real ones. The CVENET method uses a variational autoencoder, which learns a latent space representation in a probabilistic manner via an encoder and then maps that representation to the output by a decoder. 

\begin{figure}[!t]
\centering
\subfloat[]{\includegraphics[width=2.5in]{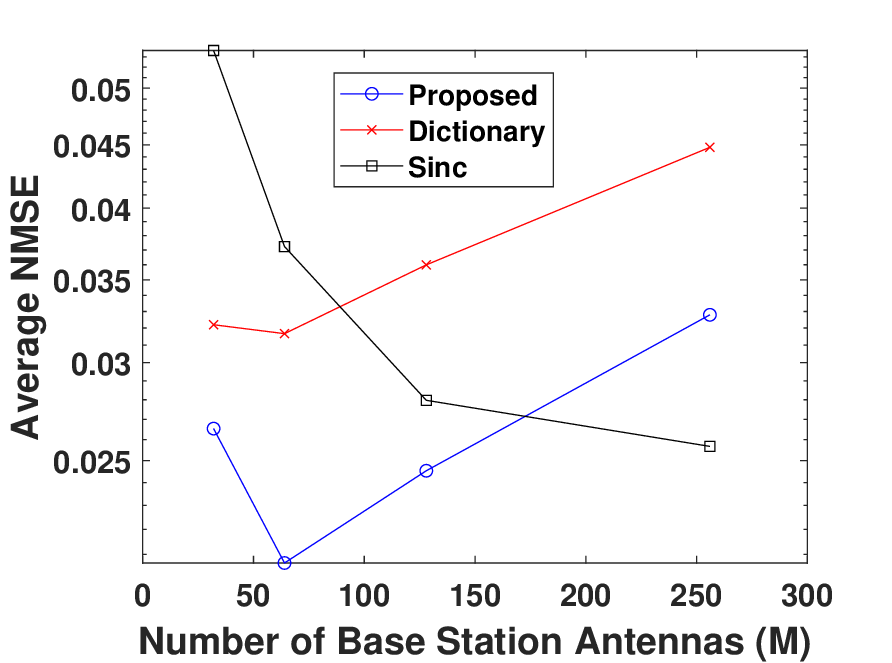}
\label{M_vs_NMSE}}
\hfil
\subfloat[]{\includegraphics[width=2.5in]{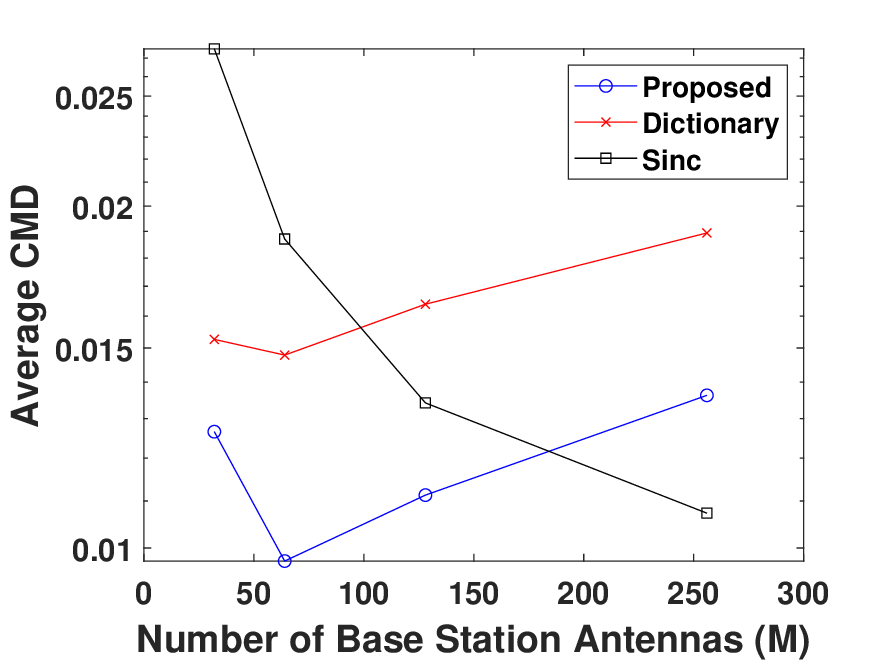}
\label{M_vs_CMD}}
\hfil
\subfloat[]{\includegraphics[width=2.5in]{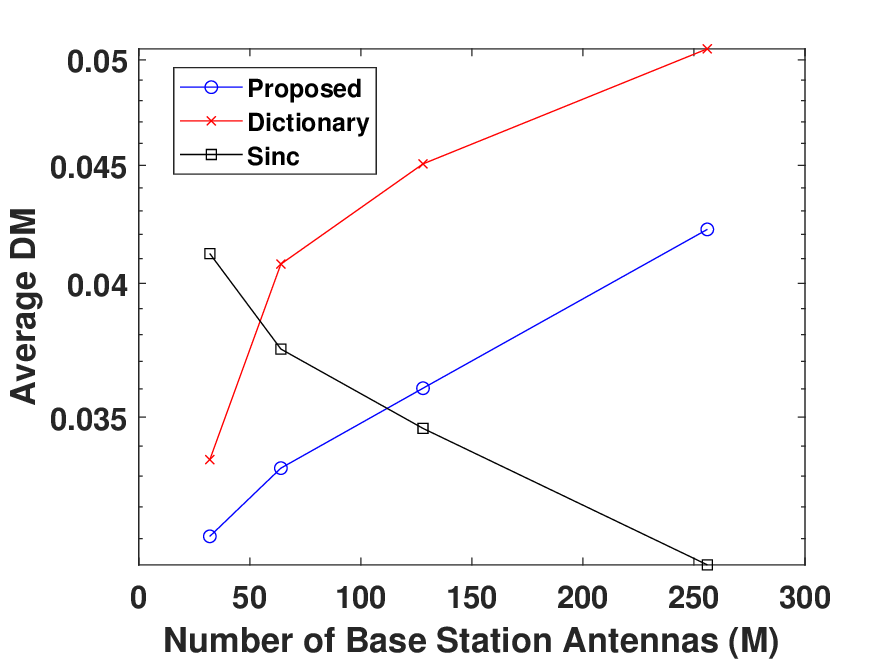}
\label{M_vs_DM}}
\caption{The variation of the errors of the compared methods with the number of base station antennas at an $\text{SNR}$ of $20 \ \text{dB}$, (a) NMSE (b) CMD (c) DM}
\label{M_vs_Error}
\end{figure}

\begin{figure}[!t]
\centering
\subfloat[]{\includegraphics[width=2.5in]{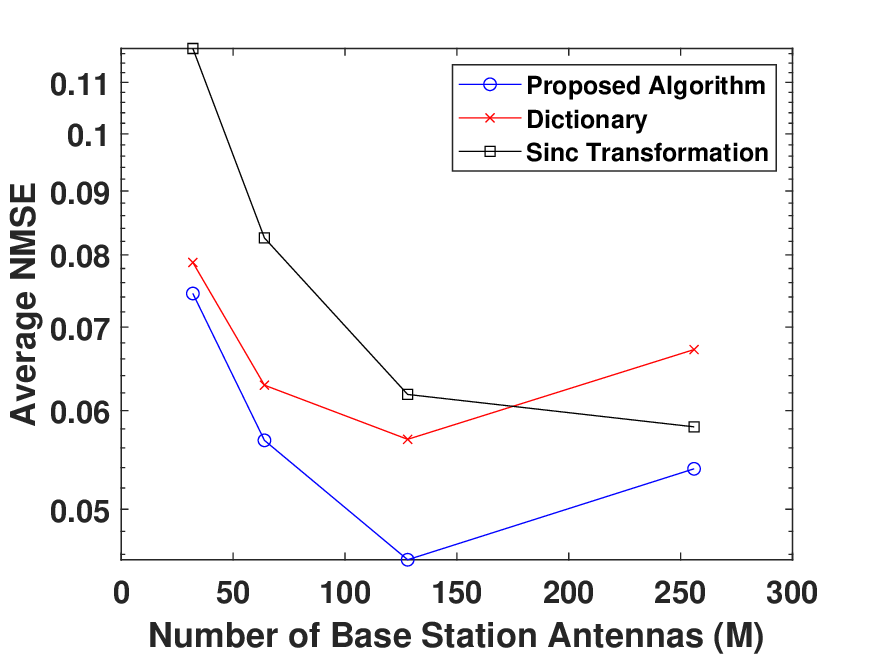}
\label{NMSE_Mover2}}
\hfil
\subfloat[]{\includegraphics[width=2.5in]{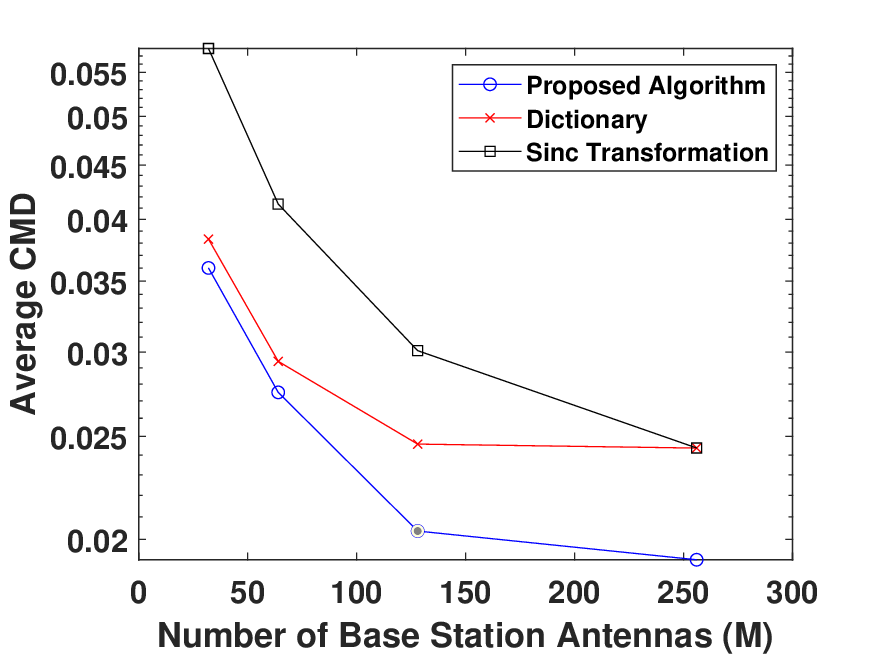}
\label{CMD_Mover2}}
\hfil
\subfloat[]{\includegraphics[width=2.5in]{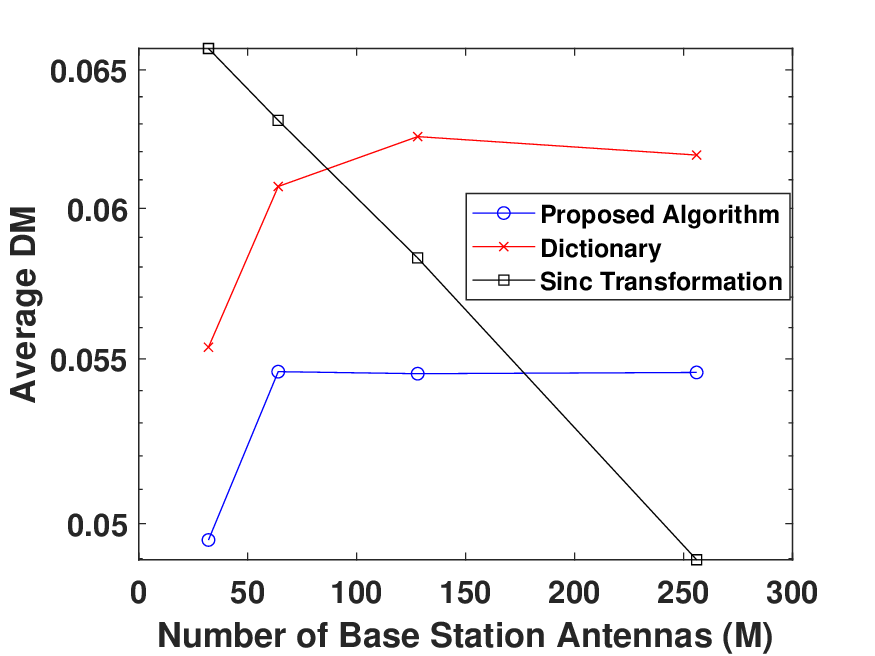}
\label{DM_Mover2}}
\caption{The variation of the errors of the compared methods with the number of base station antennas when $N_{ch}=M/2$ pilots are used, (a) NMSE (b) CMD (c) DM}
\label{M_vs_Error_Mover2}
\end{figure}

\begin{figure}[!t]
\centering
\subfloat[]{\includegraphics[width=2.5in]{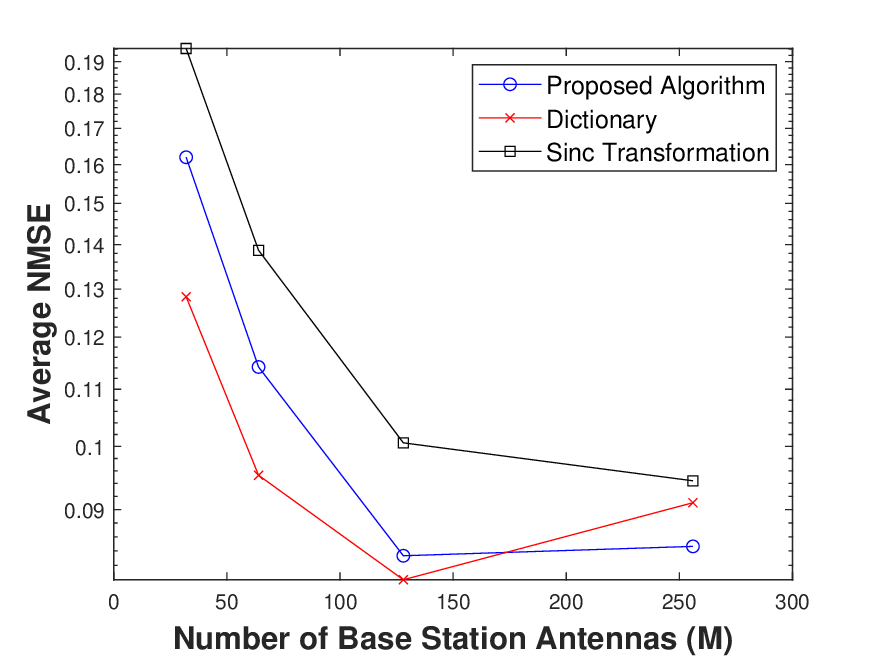}
\label{NMSE_Mover4}}
\hfil
\subfloat[]{\includegraphics[width=2.5in]{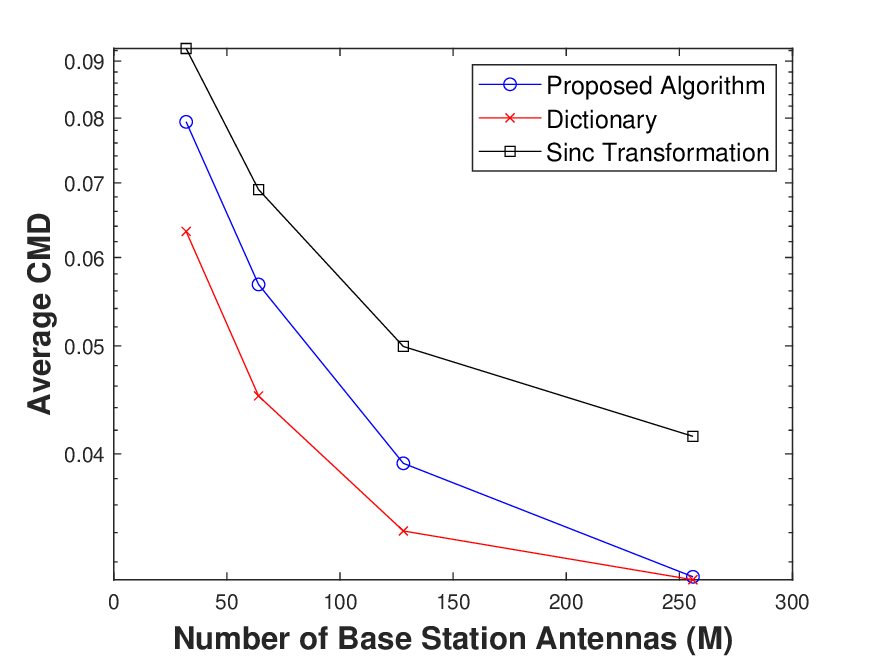}
\label{CMD_Mover4}}
\hfil
\subfloat[]{\includegraphics[width=2.5in]{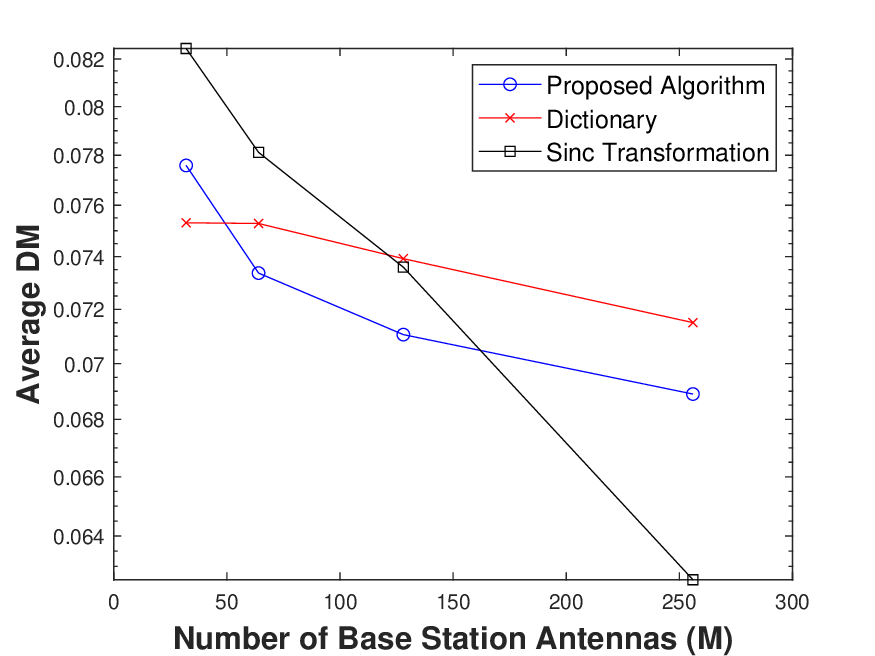}
\label{DM_Mover4}}
\caption{The variation of the errors of the compared methods with the number of base station antennas when $N_{ch}=M/4$ pilots are used, (a) NMSE (b) CMD (c) DM}
\label{M_vs_Error_Mover4}
\end{figure}

\begin{figure}[!t]
\centering
\subfloat[]{\includegraphics[width=2.5in]{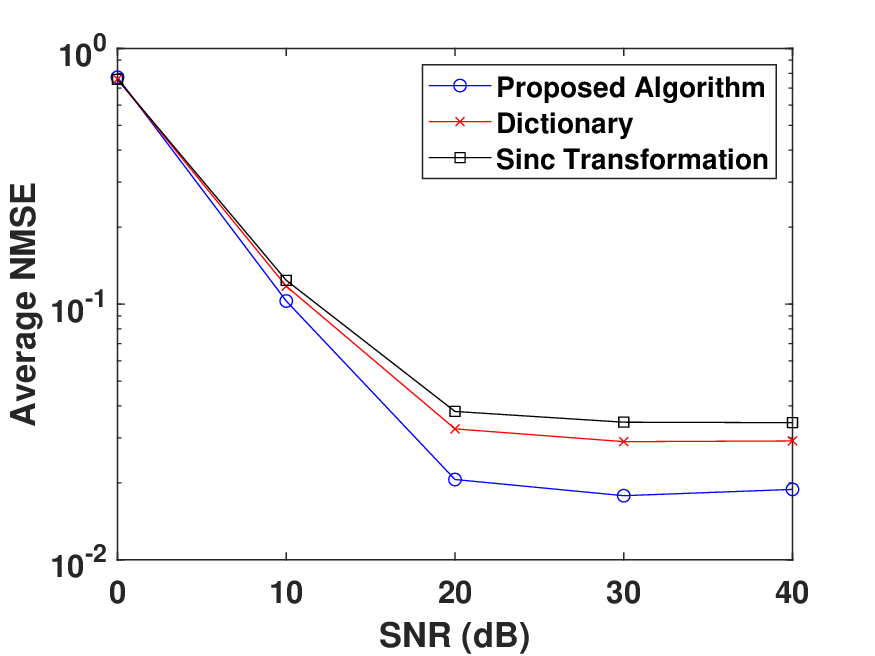}
\label{SNR_vs_NMSE}}
\hfil
\subfloat[]{\includegraphics[width=2.5in]{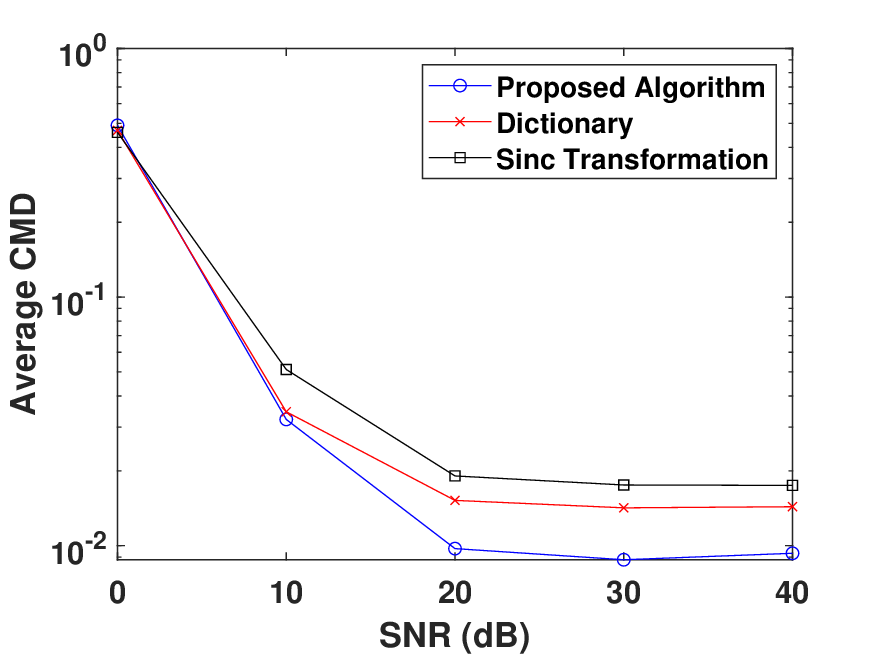}
\label{SNR_vs_CMD}}
\hfil
\subfloat[]{\includegraphics[width=2.5in]{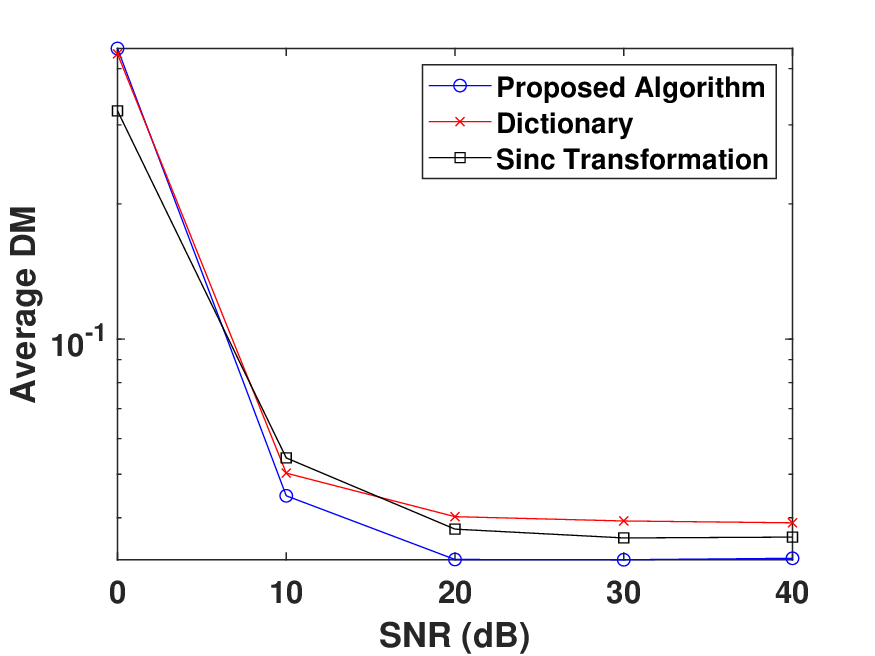}
\label{SNR_vs_DM}}
\caption{The variation of the errors of the compared methods with SNR when the number of base station antennas is $M=64$, (a) NMSE (b) CMD (c) DM}
\label{SNR_vs_Error}
\end{figure}

\begin{figure}[!t]
\centering
\subfloat[]{\includegraphics[width=2.5in]{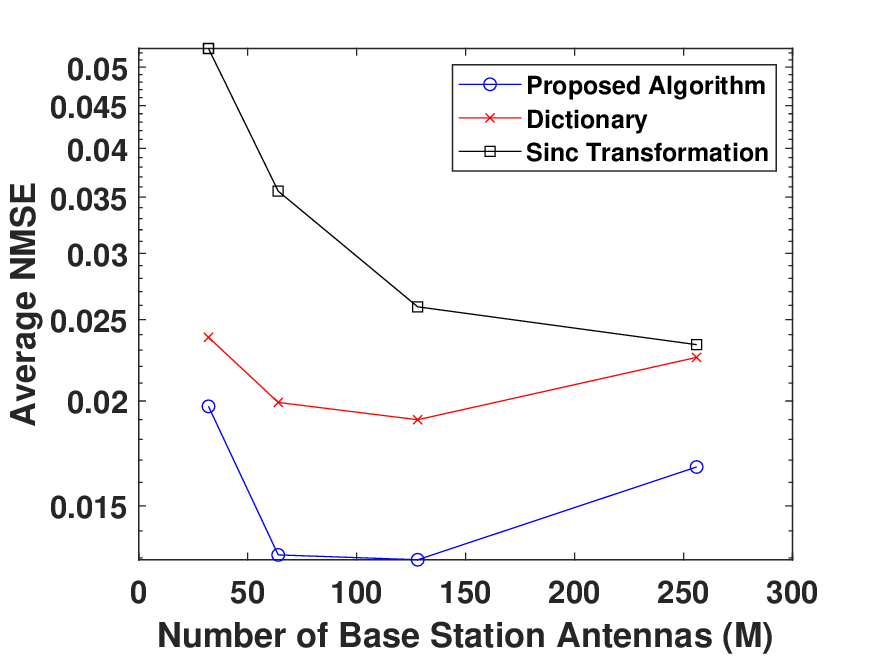}
\label{M_vs_NMSE_Lap}}
\hfil
\subfloat[]{\includegraphics[width=2.5in]{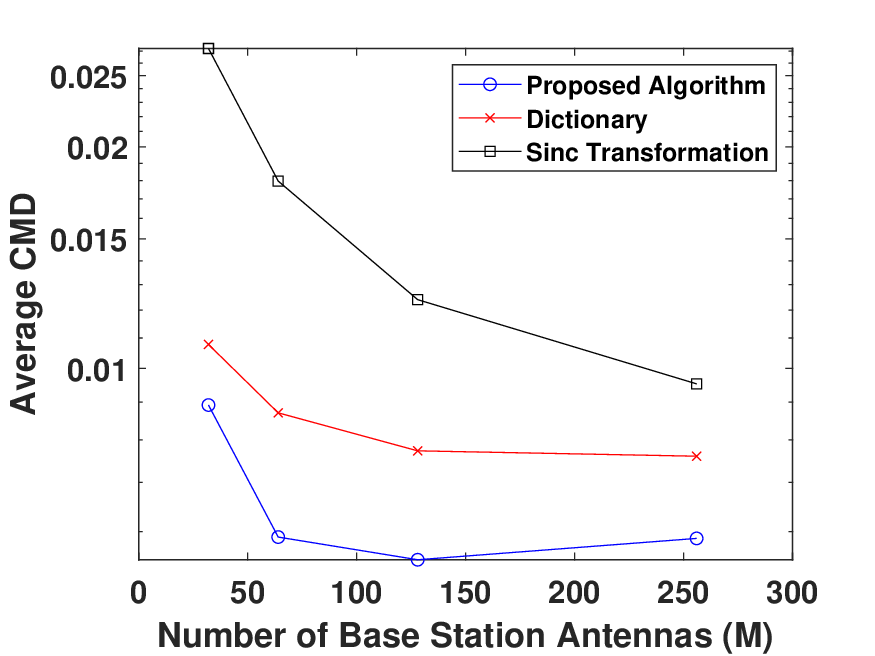}
\label{M_vs_CMD_Lap}}
\hfil
\subfloat[]{\includegraphics[width=2.5in]{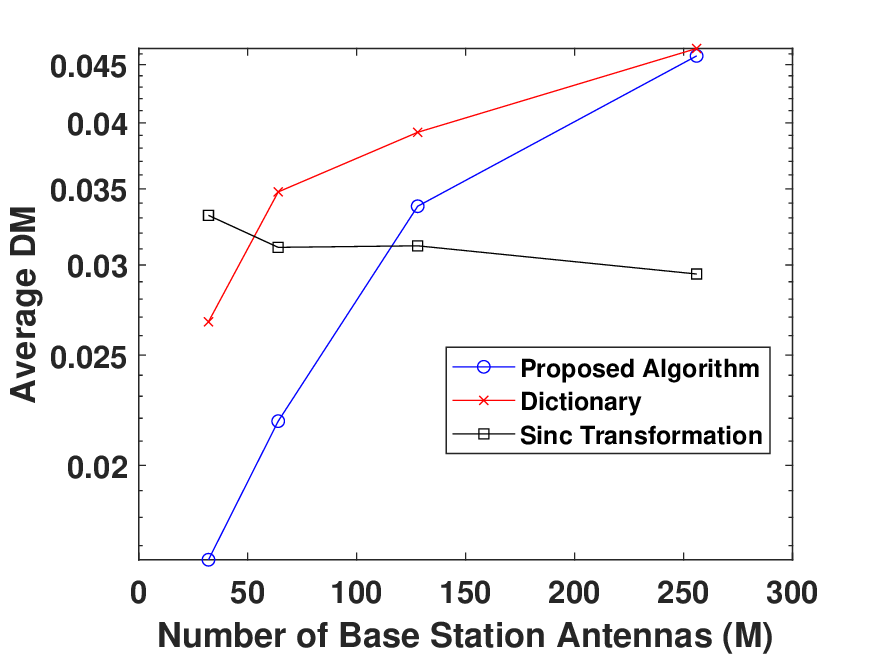}
\label{M_vs_DM_Lap}}
\caption{The variation of the errors of the compared methods with the number of base station antennas when $\text{SNR} = 20 \ \text{dB}$ and the PAS is Laplacian, (a) NMSE (b) CMD (c) DM}
\label{M_vs_Error_Lap}
\end{figure}

\begin{figure}[!t]
\centering
\subfloat[]{\includegraphics[width=2.5in]{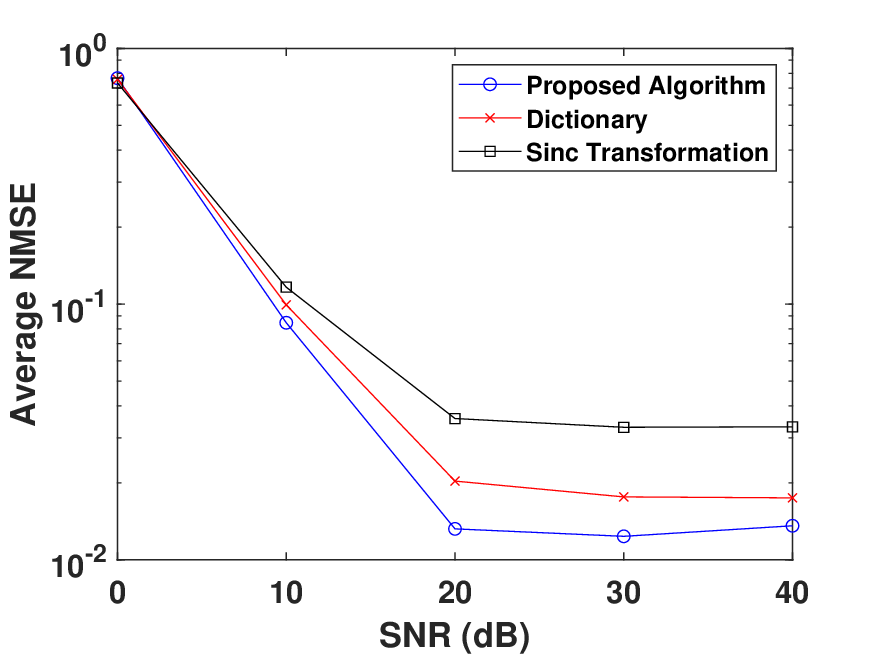}
\label{SNR_vs_NMSE_Lap}}
\hfil
\subfloat[]{\includegraphics[width=2.5in]{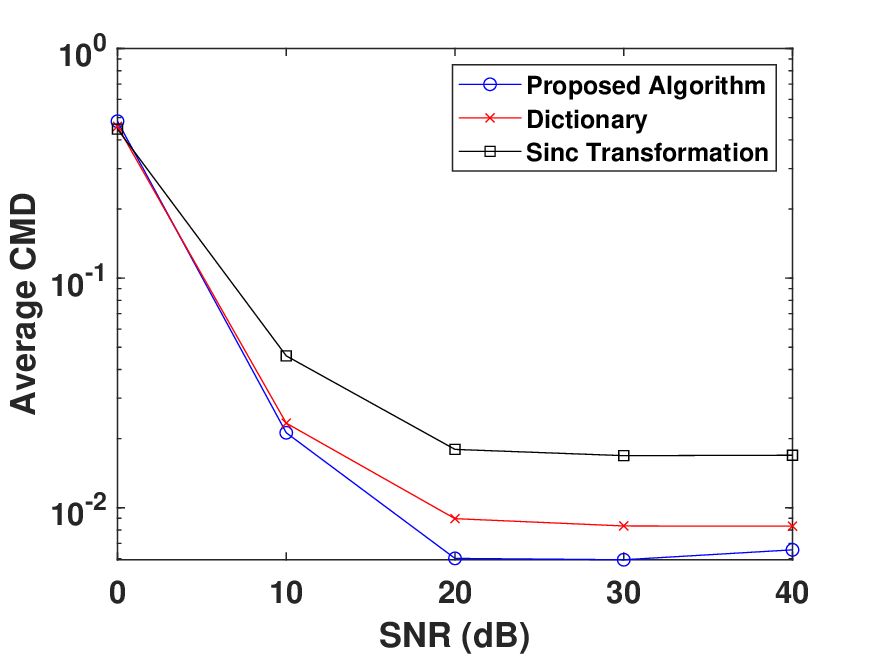}
\label{SNR_vs_CMD_Lap}}
\hfil
\subfloat[]{\includegraphics[width=2.5in]{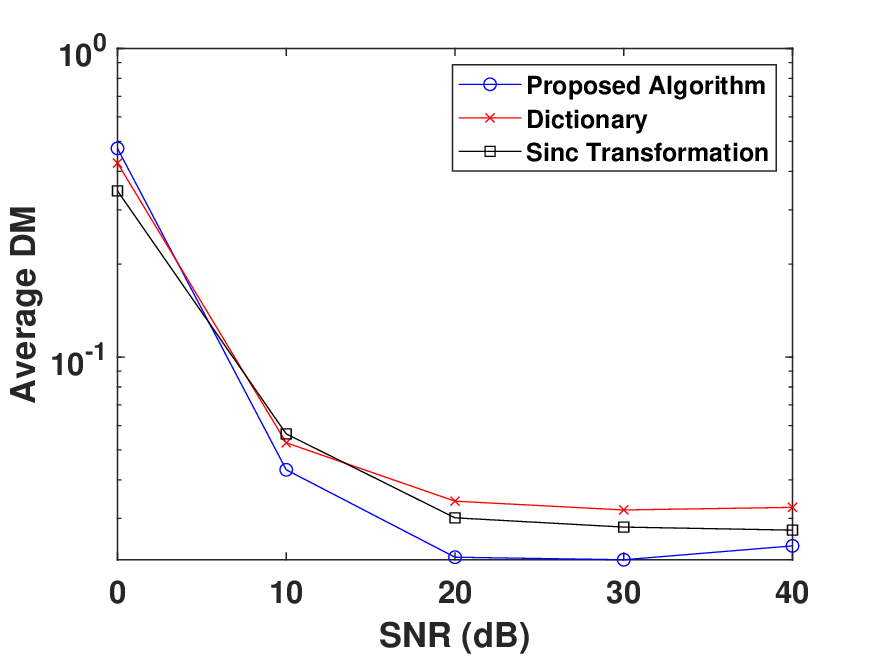}
\label{SNR_vs_DM_Lap}}
\caption{The variation of the errors of the compared methods with SNR when the number of base station antennas is $M=64$ and the PAS is Laplacian, (a) NMSE (b) CMD (c) DM}
\label{SNR_vs_Error_Lap}
\end{figure}

\begin{figure}[!t]
\centering
\subfloat[]{\includegraphics[width=2.5in]{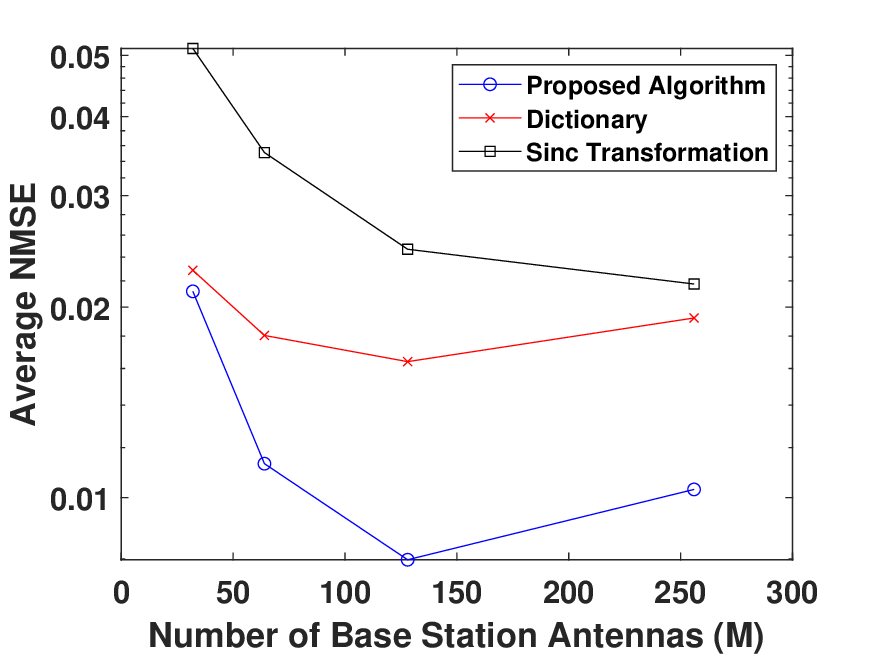}
\label{M_vs_NMSE_Gauss}}
\hfil
\subfloat[]{\includegraphics[width=2.5in]{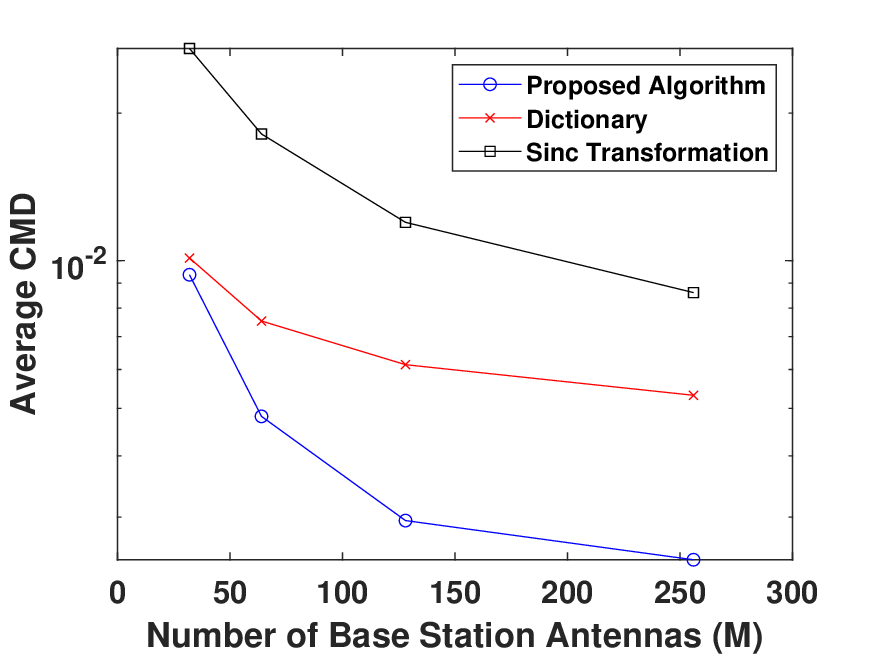}
\label{M_vs_CMD_Gauss}}
\hfil
\subfloat[]{\includegraphics[width=2.5in]{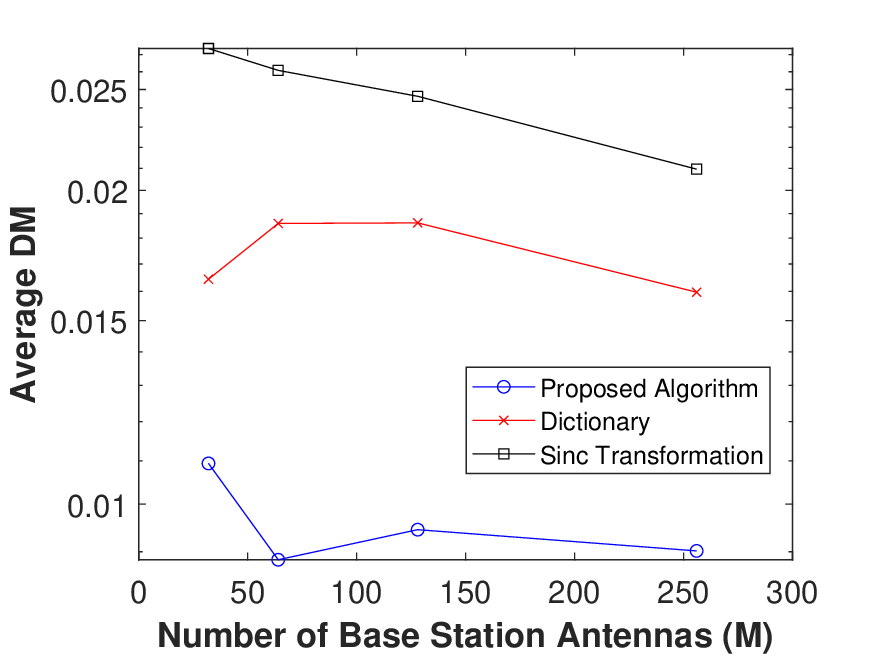}
\label{M_vs_DM_Gauss}}
\caption{The variation of the errors of the compared methods with the number of base station antennas when $\text{SNR} = 20 \ \text{dB}$ and the PAS is Gaussian, (a) NMSE (b) CMD (c) DM}
\label{M_vs_Error_Gauss}
\end{figure}

\begin{figure}[!t]
\centering
\subfloat[]{\includegraphics[width=2.5in]{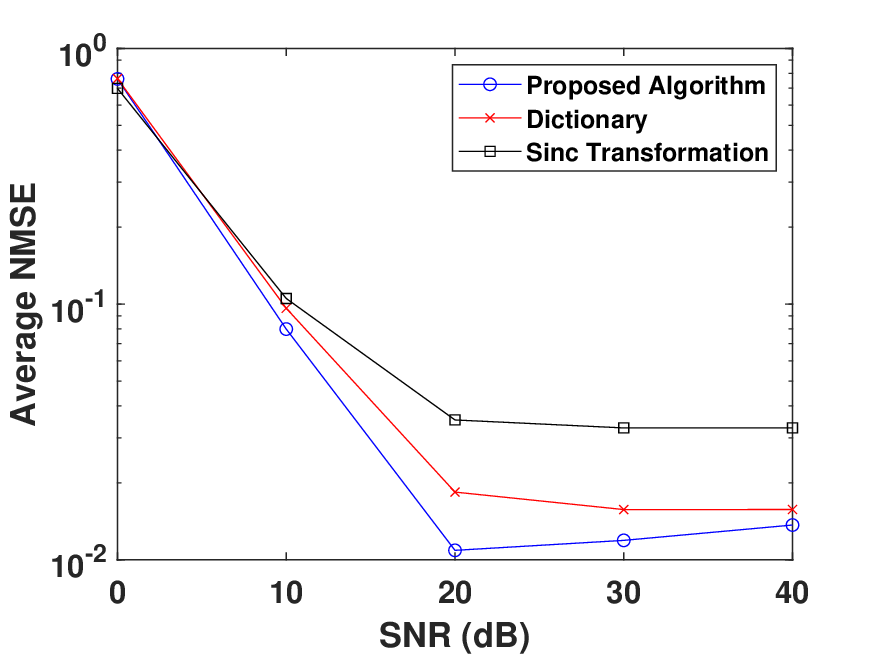}
\label{SNR_vs_NMSE_Gauss}}
\hfil
\subfloat[]{\includegraphics[width=2.5in]{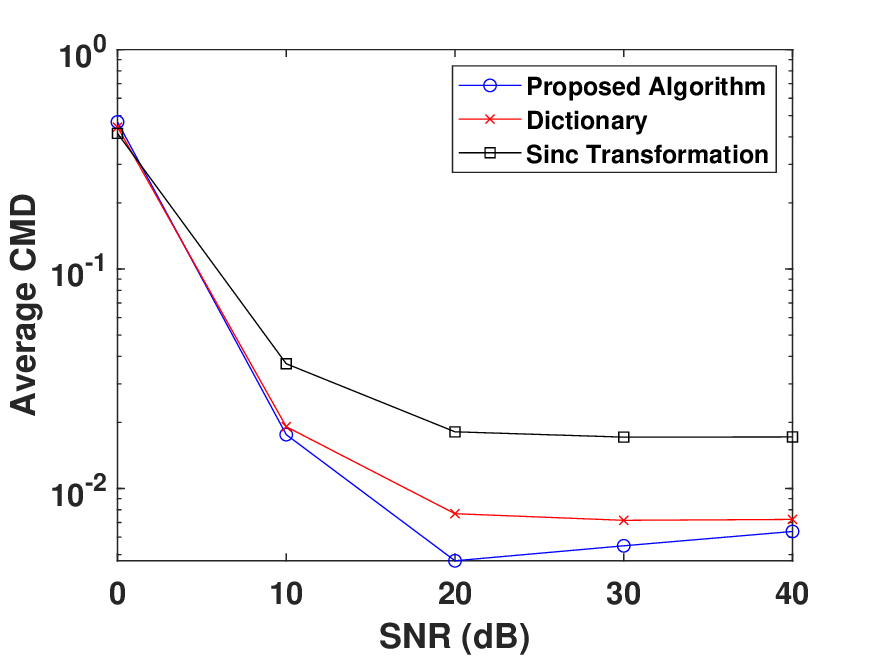}
\label{SNR_vs_CMD_Gauss}}
\hfil
\subfloat[]{\includegraphics[width=2.5in]{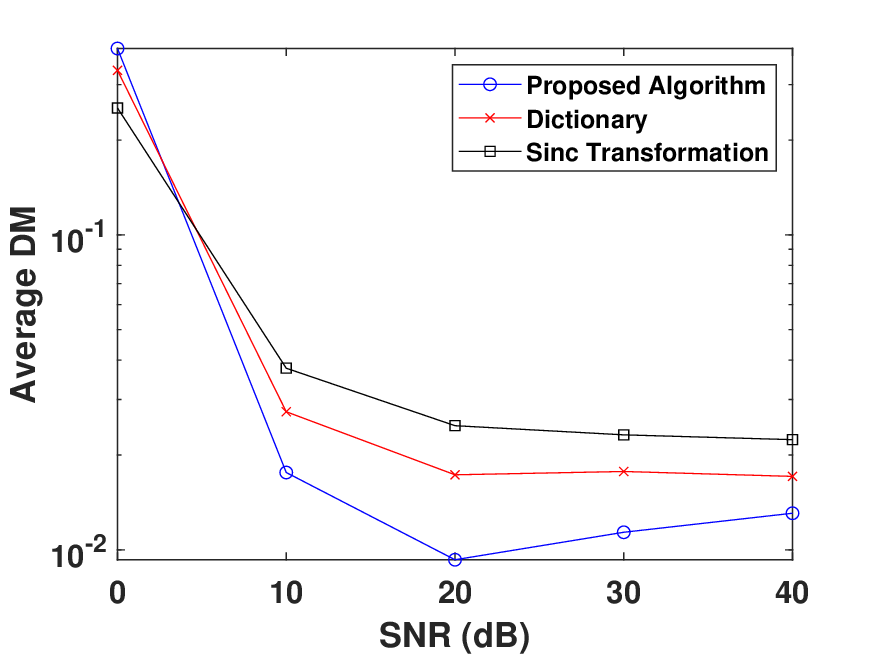}
\label{SNR_vs_DM_Gauss}}
\caption{The variation of the errors of the compared methods with SNR when the number of base station antennas is $M=64$ and the PAS is Gaussian, (a) NMSE (b) CMD (c) DM}
\label{SNR_vs_Error_Gauss}
\end{figure}

We conduct three different experiments. First, we calculate the DL CCM estimation errors with a perfect dataset in order to study the performance of the compared methods. Then, we calculate the DL CCM estimation errors for different SNR values. Finally, we compare the error values of the algorithms for different numbers of base station antennas, $M$. The hyperparameters $\mu_1, \mu_2, \mu_3$ of our method are selected over a validation data set generated independently of the test data set in our experiments.  When searching the optimal values of $\mu_1, \mu_2, \mu_3$ over the validation data set, we have applied a two-stage search strategy: In the first stage, we roughly determine a suitable range of $\mu_1, \mu_2, \mu_3$ values. Then, in the second stage we further optimize these weight values with a finer linear search. The algorithm hyperparameters have been set to the values indicated in Table \ref{sim_param_table} for the experiments in Figures \ref{N_vs_Error}-\ref{NMSE_vs_T_for_MMSE_ch_est} with noisy data. The results in Figure \ref{error_bar_all_methods} have been obtained with a slightly different choice of the hyperparameters ($\mu_1=10$, $\mu_2=3\times 10^{8}$, $\mu_3=10^7$), which have been selected separately for this particular setting with noiseless data.

In Figure \ref{error_bar_all_methods}, we compare the performance of all benchmark methods for $M=256$ base station antennas where the CCMs in both the training and the test datasets are perfectly known. The results are averaged over 10 i.i.d. datasets. One can see from Figure \ref{error_bar_all_methods} that our method outperforms the PHS interpolator in terms of all error metrics. This result demonstrates that our embedding approach proposed in Section \ref{Problem_Form_Subsection} provides higher estimation accuracy than baseline interpolators constructed using classical techniques. Also, our method mostly outperforms the dictionary method and the sinc transformation method, while the CGAN and CVENET methods have relatively higher error values than the other methods. In particular, our method yields the smallest average NMSE value of $6.1\times 10^{-3}$ among all methods, while its closest competitor algorithms dictionary and sinc transformation methods result in average NMSE values of $0.0324$ and $0.0112$, respectively. On the other hand, the average NMSE of the CGAN method for this setup is $0.0761$, and that of the CVENET method is $0.1163$.
In fact, in Figure \ref{N_vs_Error}, where we study the variation of the error with dataset size, our method achieves an average NMSE value smaller than $0.0761$ with a dataset size of only $75$, while the error of the CGAN algorithm remains above this value until the dataset size is increased to its maximal value $500$ in this experiment. One can interpret this finding as follows: Even though deep learning methods can successfully learn highly complex functions, they need a large amount of data to achieve this. 
In settings with a limited availability of training data, such methods may fail to learn a network that generalizes to new data well. Considering also the long training processes, in the rest of our experiments we compare our algorithm only with the dictionary and the sinc transformation methods, since they are closer to our algorithm in terms of performance.
From Figure \ref{N_vs_Error}, one can also conclude the following: Given that the sinc interpolation method does not rely on a dataset, its performance does not change with the dataset size, $N$. It can be seen from Figure \ref{N_vs_Error} that our method outperforms the sinc interpolation method with a dataset of size $N=500$, while the other two benchmark methods seem to need larger datasets to do this.

Figure \ref{M_vs_Error} presents the average errors obtained with the compared algorithms where the number of base station antennas varies in the range $M \in \{ 32,64,128,256\}$. For $25$ i.i.d. datasets, the experiments are repeated and the average errors are reported in Figure \ref{M_vs_Error}. One can see that the proposed algorithm outperforms the dictionary method with respect to each error metric for all numbers of antennas. However, the sinc transformation method yields smaller average error than our method when the number of antennas is high, e.g., at $M=256$. This result is expected, since both the dictionary method and our algorithm rely on training data, while estimating more matrix parameters with the same dataset size becomes increasingly challenging as the number of base station antennas grows. On the other hand, the sinc transformation method has an error upper bound that decreases with the number $M$ of antennas, as discussed in \cite{10189874}. 
Even though the average error of the sinc transformation method is lower than that of our method for $M=256$ antennas, we have observed the standard deviations of the NMSE values for our method, the dictionary method and the sinc transformation method to be 0.0161, 0.0377 and 0.0343, respectively. One can deduce from these results that although our algorithm may yield higher average error than the sinc transformation method at a high number of antennas, its performance is more stable than that of the sinc transformation method, i.e., it is less likely to exhibit erratic, excessively high error values. 

We next study the performance of our algorithm when a smaller number of pilot samples are used. In Figure \ref{M_vs_Error_Mover2} and \ref{M_vs_Error_Mover4}, 
we present the variation of the error with the number $M$ of base station antennas, where the number of pilots is fixed to $M/2$ and $M/4$, respectively. Comparing these results to those in Figure 4 obtained with $2M$ pilots, the reduction in the number of pilots is seen to lead to degradation in the performance for all methods. Meanwhile, the proposed method still yields the best performance among all methods in almost all cases with $M/2$ pilots. Our method outperforms the sinc transformation method and is competitive with the dictionary method for $M/4$ pilots. In particular, the proposed method  outperforms the dictionary method when the number of antennas is high (e.g., $M = 256$), which demonstrates its practical utility in scenarios where the pilot usage becomes problematic due to the increase in the number of antennas.

Figure \ref{SNR_vs_Error} shows the performance of the algorithms when the base station has $M=64$ antennas. The experiments are repeated for 25 i.i.d. datasets. In this scenario, the CCMs have been constructed for several different SNR values ranging from $0 \ \text{dB}$ to $40 \ \text{dB}$ and the effect of the SNR on the performance is observed. We observe that all algorithms yield high estimation error at 0 dB SNR as expected, where the CCMs are corrupted with severe noise. As the SNR increases, the estimates obtained from each algorithm improves and our algorithm outperforms the benchmark methods in all performance metrics.

In the experiments whose results are provided in Figure \ref{SNR_vs_Error}, the PAS used to create the CCM dataset is uniform.  We also examine the performance of our algorithm for the non-uniform PAS scenario in order to explore its generalizability to different PAS forms. Even though our theoretical analysis provides a rationale for the proposed method under the assumption of uniform PAS with a constant spread of AoA, it is 
informative to experimentally study the performance of our method
when these constraints are relaxed.
Figure \ref{M_vs_Error_Lap}, \ref{SNR_vs_Error_Lap}, \ref{M_vs_Error_Gauss} and \ref{SNR_vs_Error_Gauss} compare the performance of our method with the benchmarks for truncated Laplacian and truncated Gaussian PASs,
under variable number of BS antennas and variable SNR.
One can conclude from these figures that our method outperforms the benchmark methods in non-uniform PAS scenarios as well. 

\subsection{DL CSI Prediction via MMSE Channel Estimation}
An important application of channel covariance information is the channel estimation problem \cite{fang2017low,xie2018channel}. In this section, we test the performance of our algorithm in channel estimation. The DL CSI estimation is performed using the minimum mean squared error (MMSE) channel estimation method, which leverages the DL channel covariance information. 

The received signal after the pilot transmission in a Gaussian zero mean DL channel $\textbf{h}$ with  covariance matrix $\textbf{R}$, i.e., $\textbf{h} \sim \mathcal{CN}(\textbf{0},\,\textbf{R})$, is expressed as 
\begin{equation} \label{pilot_signal_time_p}
    y_{p}=\textbf{h}^{T}\textbf{x}_{p}+n_{p}, \ p \in \{1,...,N_{p}\}
\end{equation}
where $y_{p}$ is the received signal, $\textbf{h}$ is the DL channel, $\textbf{x}_{p}$ is the pilot symbol transmitted through the elements of the BS antenna array at time instant $p$, the variable $n_{p}\sim \mathcal{CN}(0,\,\sigma_{p}^{2})$ is the noise, and $N_{p}$ is the number of pilot symbols transmitted.
The received signals after pilot transmission can be expressed in the form of a matrix equation as
\begin{equation}
    \textbf{y}=\textbf{X}\textbf{h}+\textbf{n},
\end{equation}
where $\textbf{y}:=[y_{1} \ \dots \ y_{N_{p}}]^{T}$, $\textbf{X}:=[\textbf{x}_{1} \ \dots \ \textbf{x}_{N_{p}}]^{T}$ and $\textbf{n}:=[n_{1} \ \dots \ n_{N_{p}}]^{T}$.

The MMSE channel estimator for this setting is given by \cite{fang2017low}
\begin{equation} \label{MMSE_ch_est_formula}
    \hat{\textbf{h}}_{MMSE}=\textbf{R}\textbf{X}^{H}\left( \textbf{X}\textbf{R}\textbf{X}^{H}+\sigma_{p}^{2}\textbf{I} \right)^{-1}\textbf{y},
\end{equation}

whose mean squared error (MSE) is obtained in closed form as \cite{fang2017low}
\begin{equation} \label{MSE_of_MMSE_ch_est}
    \text{MSE}=tr \left(\textbf{R}-\textbf{R}\textbf{X}^{H}\left( \textbf{X}\textbf{R}\textbf{X}^{H}+\sigma_{p}^{2}\textbf{I} \right)^{-1} \textbf{X}\textbf{R} \right).
\end{equation}

A total pilot power constraint is employed for the pilot transmission scheme, which is $tr(\textbf{X}\textbf{X}^{H}) \leq P$, where $P$ is the total power allocated for the pilot symbols.

The pilot matrix $\textbf{X}$ is formed such that it has orthonormal rows, i.e., 
\begin{equation}
\textbf{x}_{i}^{H}\textbf{x}_{j}=
    \begin{cases}
        1 & \text{if $i=j$ }\\
        0 & \text{if $i \neq j$}
    \end{cases}, i,j \in \{1,...,N_{p}\}.
\end{equation}
It is then scaled to have $tr(\textbf{X}\textbf{X}^{H}) = P$ so that it obeys the power constraint rule provided above. 

We conduct experiments for $M=64$ BS antennas, where a CCM dataset is generated with an SNR of $20 \ \text{dB}$. After learning DL CCMs from their UL counterparts via the algorithms examined, we use the DL CCM estimates in the MMSE channel estimator given in \eqref{MMSE_ch_est_formula} instead of the true DL CCM values. The imperfect MMSE channel estimate of a DL channel realization $\textbf{h}$ whose true CCM is $\textbf{R}$ is given by
\begin{equation} \label{imp_MMSE_ch_est_formula}
    \hat{\textbf{h}}_{MMSE}^{imp}=
    \hat{\textbf{R}}\textbf{X}^{H}\left( \textbf{X}\hat{\textbf{R}}\textbf{X}^{H}+\sigma_{p}^{2}\textbf{I} \right)^{-1}\textbf{y},
\end{equation}
where $\hat{\textbf{h}}_{MMSE}^{imp}$ is the MMSE channel estimation of $\textbf{h}$ obtained from the estimate $\hat{\textbf{R}}$ of the true CCM $\textbf{R}$.
 The estimate $\hat{\textbf{R}}$ is obtained from 
the algorithms in comparison.

We measure the normalized channel estimation errors given by the imperfect CCM estimates through the NMSE metric defined as

\begin{equation} \label{NMSE_for_MMSE}
    \text{NMSE} = \mathop{\mathbb{E}} \Biggl\{\frac{\|\textbf{h}-\hat{\textbf{h}}\|^2}{\|\textbf{h}\|^2}\Biggr\},  
\end{equation}

where $\textbf{h}$ is the true value of a DL channel realization and $\hat{\textbf{h}}$ is the MMSE channel estimate found through the DL CCM.

We conduct the experiment on the same 25 i.i.d. datasets used in the previous experiments where the number of BS antennas is $M=64$ and the SNR is $20 \ \text{dB}$. In order to compare the performance of the imperfect MMSE channel estimator \eqref{imp_MMSE_ch_est_formula} with the tested CCM estimation methods to that of MMSE channel estimation with the true DL CCM, the following procedure is applied:

\begin{enumerate}
    \item First, for each dataset, the DL CCMs are estimated for the test points. 
    \item Using the true DL CCM $\textbf{R}$ of each test point, 100 different DL channel realizations are constructed as 
    \begin{equation} \label{channel_realizations_MMSE}      \textbf{h}_{r}=\textbf{R}^{1/2}
        \textbf{w}_{r}, \ r=1,...,100,  
    \end{equation}
    where $\textbf{w}_{r} \sim \mathcal{CN}(\textbf{0},\,\textbf{I})$.
    \item Next, for each channel realization given by \eqref{channel_realizations_MMSE}, the MMSE estimate $\hat{\textbf{h}}_{r}$ is computed using \eqref{MMSE_ch_est_formula}.
     \item Then, using the DL CCM estimates $\hat{\textbf{R}}$ found with each DL CCM estimation method, the imperfect MMSE channel estimates $\hat{\textbf{h}}_{r}^{imp}$ are calculated using \eqref{imp_MMSE_ch_est_formula} for all 100 channel realizations for each method.
     \item  Finally, the NMSE is calculated via \eqref{NMSE_for_MMSE} for $\hat{\textbf{h}}_{r}$ and for the estimates $\hat{\textbf{h}}_{r}^{imp}$ of the DL CCM methods.
\end{enumerate}

Each of the 25 datasets contains 100 test points, with 100 channel realizations for each test point. The expectation in the NMSE expression \eqref{NMSE_for_MMSE} is computed numerically by averaging first over the 100 channel realizations, then over the 100 test points, and finally over the 25 test datasets.

\begin{figure}[!t]
\centering
\includegraphics[width=2.5in]{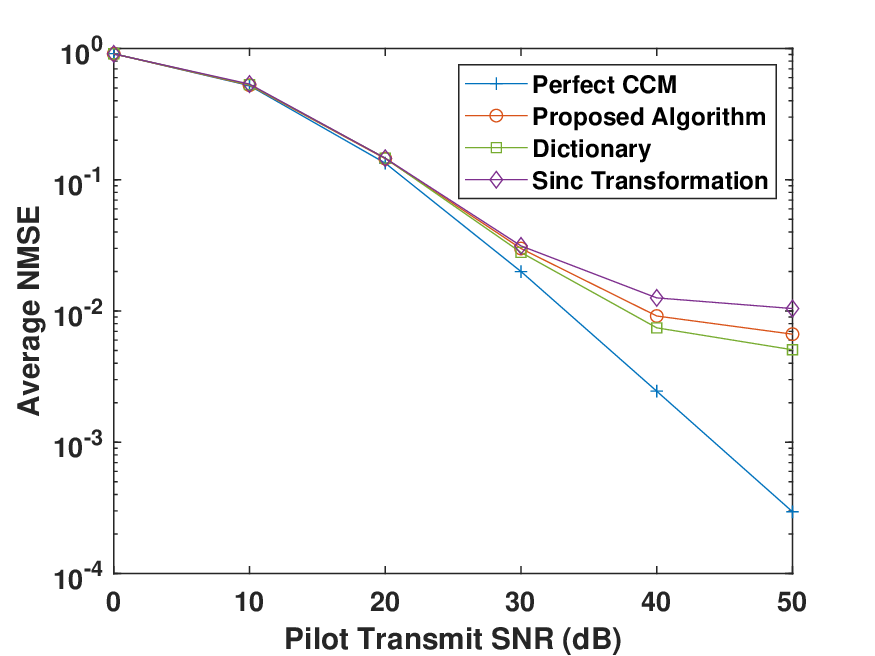}
\caption{The variation of the average NMSE of the MMSE channel estimator with the pilot transmit SNR}
\label{NMSE_vs_SNR_for_MMSE_ch_est}
\end{figure}

\begin{figure}[!t]
\centering
\includegraphics[width=2.5in]{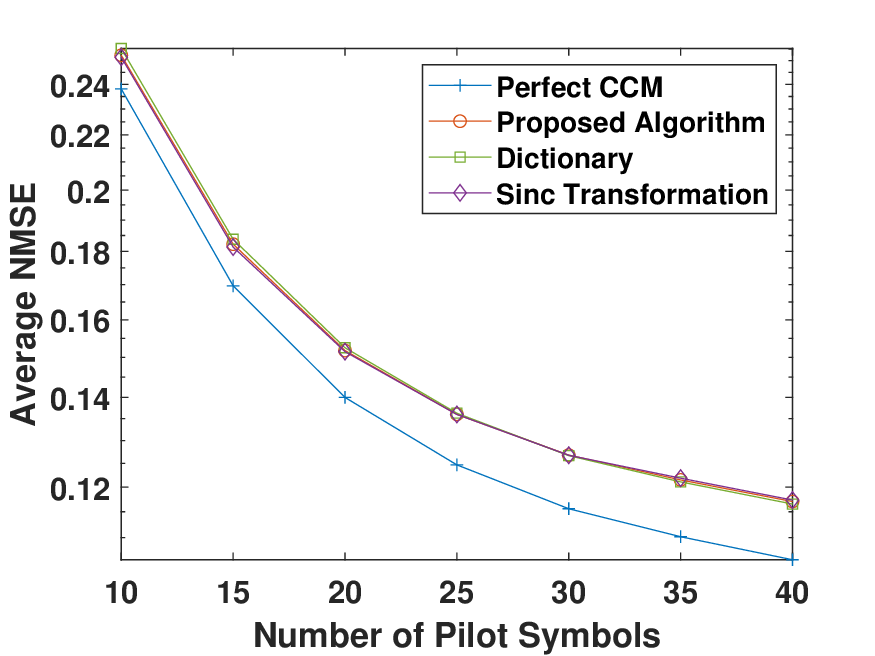}
\caption{The variation of the average NMSE of the MMSE channel estimator with the number of pilot symbols}
\label{NMSE_vs_T_for_MMSE_ch_est}
\end{figure}

Two different experiments are conducted. The first setup studies the effect of the SNR in pilot signaling on the channel estimation performance. The pilot transmit power is set to $P$ and the noise power is given by $\sigma_{p}^2$ as explained previously.
Thus, we can define the pilot transmit SNR as $P / \sigma_{p}^2$. The MMSE channel estimators with the perfect CCM and with the CCM estimates are compared in terms of NMSE in Figure \ref{NMSE_vs_SNR_for_MMSE_ch_est} for pilot transmit SNR values between $0 \ \text{dB}$ and $50 \ \text{dB}$, and for a constant pilot signaling time chosen as the rank of the true DL CCM value. The second experiment investigates the effect of the number of pilot symbols on channel estimation, whose results are presented in Figure \ref{NMSE_vs_T_for_MMSE_ch_est}. The number of pilot symbols ranges from 10 to 40, and the pilot transmit SNR is set to $20 \ \text{dB}$ for this setting.

\FloatBarrier

\FloatBarrier

In Figure \ref{NMSE_vs_SNR_for_MMSE_ch_est}, we observe that the error of the MMSE channel estimate with perfect CCM approaches zero, while the errors of the MMSE channel estimates with the imperfect CCMs saturate, as the pilot transmit SNR increases.
This is not a surprising result, since as the pilot transmit SNR increases, the power of noise in the received signals after pilot signaling approaches zero.
This suggests that the MMSE channel estimate using the true DL CCM approaches the true channel realization considering that there is no noise in the channel observations and the CCM of the channel is known perfectly. Also, the number of pilot symbols is chosen as equal to the rank of the CCM, which allows the MSE in \eqref{MSE_of_MMSE_ch_est} to approach zero with diminishing noise \cite{fang2017low}. On the other hand, even though the channel observations become noise-free for the imperfect CCMs as well, this does not remove the imperfections in the CCM estimates. Therefore, due to these residual errors, the improvement in the performance of imperfect MMSE channel estimates starts to slow down as the pilot transmit SNR increases, especially at high values where the noise could be considered as almost zero.

If we compare the MMSE channel estimation performance of the compared methods,
the following can be concluded from Figure \ref{NMSE_vs_SNR_for_MMSE_ch_est}: As one can see in Figure \ref{SNR_vs_Error}, our DL CCM estimation algorithm outperforms the benchmark methods in terms of all three error metrics. However, there is no direct parallel between the estimation accuracy of the DL CCM and the performance of the imperfect MMSE channel estimation employing it. In fact, at high pilot transmit SNR values, the performance of the MMSE estimate of the dictionary method  
slightly surpasses that of our algorithm. This result may seem surprising at first; however, it could be  due to the fact that the design of the DL CCM estimation algorithms do not explicitly aim to minimize the MSE of the MMSE channel estimator. Considering that the dictionary method offers a solution where the DL CCM estimate is a weighted average of the DL CCMs of the other user points in the dictionary, the resulting interpolation provides a viable DL CCM estimate. Note that our method and the sinc transformation method rely on the estimation of the first row of the DL CCM rather than the whole matrix.
Due to the reason above, the dictionary method may provide DL CCM estimates
with slightly
stronger structural integrity, resulting in marginally
better MMSE channel estimates even though it provides worse CCM estimates in terms of the error metrics.
Overall, the MMSE channel estimation error of our method is quite close to that of the dictionary method, the maximal MMSE gap between them being around $10^{-3}$.

The results in Figure \ref{NMSE_vs_T_for_MMSE_ch_est} similarly indicate that, although our algorithm outperforms the other methods in terms of DL CCM estimation, the performance gap between them almost vanishes in MMSE channel estimation, as in the case of constant pilot time and variable pilot transmit SNR.

An overall consideration of our experimental results suggests that the proposed algorithm performs
better than the other methods in terms of DL CCM estimation and yields very similar results in terms of MMSE channel estimation.
This shows that our algorithm has the potential to be useful in this application area. In order to achieve better MMSE channel estimates, 
the objective function \eqref{Overall_Problem} of our method may be extended to incorporate an additional term representing the MSE of the MMSE channel estimate, which is an interesting future direction of our study.

\begin{remark}
    \normalfont
    While the dictionary method yields marginally better MMSE channel estimates than the proposed method,
    its computational complexity is given as $\mathcal{O}(M^4N)$ in \cite{decurninge2015channel}. 
    Recalling that our method has a much smaller computational complexity of $\mathcal{O}(M^2N)$ in the test phase, one may conclude that the proposed method provides a quite favorable performance-complexity tradeoff especially in settings with a high number of base station antennas, such as massive MIMO systems.
\end{remark}

\section{Conclusion} \label{Conclusion_Section}
In this paper, we have proposed a novel DL CCM estimation method for FDD massive MIMO systems where the base station is equipped with ULA antennas. We have first presented a theoretical analysis that gives an upper bound on the estimation error of the DL CCM from UL CCMs. We have then proposed a representation learning method that constructs an analytical mapping from UL CCMs to their DL CCM counterparts. The proposed method aims at learning an interpolation function from datasets relatively smaller than those needed for training deep neural networks, while benefiting from the richness of the underlying nonlinear model so that the resulting mapping is more robust to variations in the system parameters than simple signal processing solutions. Experimental results show that the proposed algorithm achieves better estimation performance than the benchmark methods in most of the scenarios. 
The proposed method can especially be useful in practical applications with limited access to training data. Our algorithm shows promising performance in such applications as it provides quite accurate downlink channel covariance estimates with a simple nonlinear learning setup. The extension of our method to other base station antenna structures such as  uniform rectangular array (URA), and to other communication settings, such as multi-cluster massive MIMO scenarios, are left as future research directions. 

\section*{Acknowledgments}
The help of Dr. -Ing. Bitan Banerjee during the implementation of the CGAN-based benchmark method in \cite{banerjee2022downlink} is acknowledged.
This work was supported by Turkcell Technology under
5G and Beyond Joint Graduate Support Program run by the Information and Communication Technologies Authority of Türkiye.

{\appendix

\section{\textbf{Approximations Used in the Proof of Theorem ~\ref{lemma-2}}} \label{Appendix_theorem1_approximations}
Based on the first-order Taylor approximations given in \eqref{taylor_app_plus} and \eqref{taylor_app_minus}, one can obtain the following approximate expressions, which are useful for the proof of Theorem \ref{lemma-2}: 

\begin{eqnarray}
\frac{\sin(\bar{v_i}+\Delta)+\sin(\bar{v_j}+\Delta)}{2} \approx \sin\left(\frac{\bar{v_i}+\bar{v_j}}{2}+\Delta\right) \label{appAeqn1} \\
\frac{\sin(\bar{v_i}-\Delta)+\sin(\bar{v_j}-\Delta)}{2} \approx \sin\left(\frac{\bar{v_i}+\bar{v_j}}{2}-\Delta\right) \label{appAeqn2} \\
\pi \left(\sin(\bar{v_i}+\Delta)-\sin(\bar{v_j}+\Delta)\right) \approx \alpha_{1} (\bar{v_i}-\bar{v_j}) \label{appAeqn3} \\
\pi \left(\sin(\bar{v_i}-\Delta)-\sin(\bar{v_j}-\Delta)\right) \approx \alpha_{2} (\bar{v_i}-\bar{v_j}) \label{appAeqn4}
\end{eqnarray}

\section{\textbf{Proof of Theorem ~\ref{lemma-2}}} \label{Appendix_theorem_1_proof}
\begin{proof}  
The square of the norm of the difference between ${\textbf{p}_{UL}}^{i} \in \mathbb{R}^{1 \times 2M-1}$ and ${\textbf{p}_{UL}}^{j} \in \mathbb{R}^{1 \times 2M-1}$, which are drawn i.i.d. from $\upsilon$, is given by

\begin{multline} \label{lemma_2_start_eqn}
    \left\|{\textbf{p}_{UL}}^{i}-{\textbf{p}_{UL}}^{j} \right\|^2 = \sum_{m=1}^{M} \left|{[{\textbf{p}_{UL}}^i]}_m-{[{\textbf{p}_{UL}}^j]}_m \right|^2 \\
     =  \sum_{m=1}^{M} \Bigg| \int_{\bar{v_i}-\Delta}^{\bar{v_i}+\Delta} \frac{1}{2\Delta} \exp\left(j \pi (m-1)\sin(\phi)\right) \,d\phi \\
     - \int_{\bar{v_j}-\Delta}^{\bar{v_j}+\Delta} \frac{1}{2\Delta} \exp\left(j \pi (m-1)\sin(\phi)\right) \,d\phi \Bigg|^2 \\
     =  \sum_{m=1}^{M} \Bigg| \int_{\bar{v_j}+\Delta}^{\bar{v_i}+\Delta} \frac{1}{2\Delta} \exp\left(j \pi (m-1)\sin(\phi)\right) \,d\phi \\
     - \int_{\bar{v_j}-\Delta}^{\bar{v_i}-\Delta} \frac{1}{2\Delta} \exp\left(j \pi (m-1)\sin(\phi)\right) \,d\phi \Bigg|^2.
\end{multline}
Let us define $\theta := \pi \sin(\phi)$. The limits of the integrals in \eqref{lemma_2_start_eqn} cover very narrow intervals due to the condition $\bar{v_i}-\bar{v_j}\approx0$ considered in Theorem \ref{lemma-2}. Therefore, one can approximate $\theta$ as a linear function of $\phi$ within these intervals using a first order Taylor approximation. 
Let $\bar{v}_{\Delta}^{+}:=\left(\frac{\bar{v_i}+\bar{v_j}}{2}+\Delta\right)$ and $\bar{v}_{\Delta}^{-}:=\left(\frac{\bar{v_i}+\bar{v_j}}{2}-\Delta\right)$.
For $\phi \in \left[\bar{v_j}+\Delta,\bar{v_i}+\Delta\right]$, 
\begin{equation} \label{taylor_app_plus}
    \sin(\phi) \approx \sin\left(\bar{v}_{\Delta}^{+}\right)
    +\left(\cos\left(\bar{v}_{\Delta}^{+}\right)\right)\left(\phi-\bar{v}_{\Delta}^{+}\right). 
\end{equation}\\
Therefore, one can approximate $\theta$ as $\theta \approx \alpha_{1}\phi+\beta_{1}$, \\
where 
\begin{equation} \alpha_{1} = \pi \cos\left(\bar{v}_{\Delta}^{+}\right)
\end{equation}\\
and 
\begin{equation} 
\beta_{1} = \pi \Bigg[\sin\left(\bar{v}_{\Delta}^{+}\right)
-\bar{v}_{\Delta}^{+} \cos\left(\bar{v}_{\Delta}^{+}\right) \Bigg]. 
\end{equation} 
Similarly,
for $\phi \in \left[\bar{v_j}-\Delta,\bar{v_i}-\Delta\right]$, 
\begin{equation} \label{taylor_app_minus}
\sin(\phi) \approx \sin\left(\bar{v}_{\Delta}^{-}\right)
+\left(\cos\left(\bar{v}_{\Delta}^{-}\right)\right)\left(\phi-\bar{v}_{\Delta}^{-}\right). 
\end{equation} \\
Hence, one can approximate $\theta$ as $\theta \approx \alpha_{2}\phi+\beta_{2}$, \\
where 
\begin{equation} \alpha_{2} = \pi \cos\left(\bar{v}_{\Delta}^{-}\right) 
\end{equation} \\
and 
\begin{equation} \beta_{2} = \pi \Bigg[\sin\left(\bar{v}_{\Delta}^{-}\right)
-\bar{v}_{\Delta}^{-} \cos\left(\bar{v}_{\Delta}^{-}\right)\Bigg]. 
\end{equation} \\

Following the change of variables and approximations above, one can write $ \left| [{\textbf{p}_{UL}}^{i}]_m-[{\textbf{p}_{UL}}^{j}]_m \right|^2$ as 
\begin{multline}
        \left| [{\textbf{p}_{UL}}^{i}]_m-[{\textbf{p}_{UL}}^{j}]_m \right|^2 \\
     \approx \Bigg| \int_{\pi \sin(\bar{v_j}+\Delta)}^{\pi \sin(\bar{v_i}+\Delta)} \frac{\exp\left(j(m-1)\theta\right)}{2\Delta\alpha_{1}}  \,d\theta
     - \int_{\pi \sin(\bar{v_j}-\Delta)}^{\pi \sin(\bar{v_i}-\Delta)} \frac{\exp\left(j(m-1)\theta\right)}{2\Delta\alpha_{2}}  \,d\theta \Bigg|^2 
\end{multline}
\begin{multline} \label{eqnbeforeAppAapprox}
= \Bigg| \frac{\exp\left(j(m-1)\pi \frac{\sin(\bar{v_i}+\Delta)+\sin(\bar{v_j}+\Delta)}{2}\right)}{\Delta (m-1)\alpha_{1}} 
\\
\sin\left((m-1)\pi \frac{\sin(\bar{v_i}+\Delta)-\sin(\bar{v_j}+\Delta)}{2}\right) \\
- \frac{\exp\left(j(m-1)\pi \frac{\sin(\bar{v_i}-\Delta)+\sin(\bar{v_j}-\Delta)}{2}\right)}{\Delta (m-1)\alpha_{2}} 
\\
\sin\left((m-1)\pi \frac{\sin(\bar{v_i}-\Delta)-\sin(\bar{v_j}-\Delta)}{2}\right) \Bigg|^2.
\end{multline}

From the relations \eqref{appAeqn1}, \eqref{appAeqn2}, \eqref{appAeqn3} and \eqref{appAeqn4} provided in \ref{Appendix_theorem1_approximations}, one can approximate the expression in \eqref{eqnbeforeAppAapprox} as 

\begin{multline} \label{before_sin_taylor_exp}
\Bigg| \frac{\exp\left(j(m-1)\pi \sin\left(\bar{v}_{\Delta}^{+}\right)\right)}{\Delta (m-1)\alpha_{1}} 
\sin\left((m-1) \frac{\alpha_{1}(\bar{v_i}-\bar{v_j})}{2}\right) \\
   - \frac{\exp\left(j(m-1)\pi \sin\left(\bar{v}_{\Delta}^{-}\right)\right)}{\Delta (m-1)\alpha_{2}} 
   \sin\left((m-1) \frac{\alpha_{2}(\bar{v_i}-\bar{v_j})}{2}\right) \Bigg|^2. 
\end{multline}
Note that for $\bar{v_i}-\bar{v_j} \approx 0 $, from the first-order Taylor expansion we obtain $\sin\left((m-1)\frac{\alpha_{1}(\bar{v_i}-\bar{v_j})}{2}\right) \approx (m-1) \frac{\alpha_{1}(\bar{v_i}-\bar{v_j})}{2} $ and $\sin\left((m-1) \frac{\alpha_{2}(\bar{v_i}-\bar{v_j})}{2}\right) \approx (m-1) \frac{\alpha_{2}(\bar{v_i}-\bar{v_j})}{2} $ for all $m \in \{1,...,M\}$. The expression in \eqref{before_sin_taylor_exp} can then be approximated as
 \begin{equation*}
 \left(\frac{\bar{v_i}-\bar{v_j}}{2\Delta}\right)^2 
  \Bigg| \exp\left(j(m-1)\pi \sin\left(\bar{v}_{\Delta}^{+}\right)\right) 
 - \exp\left(j(m-1)\pi \sin\left(\bar{v}_{\Delta}^{-}\right)\right) \Bigg|^2 
 \end{equation*}
 \begin{multline*}
     = \left(\frac{\bar{v_i}-\bar{v_j}}{2\Delta}\right)^2  \Bigg| \exp\bigg[\frac{j(m-1)\pi}{2} \big(\sin\left(\bar{v}_{\Delta}^{+}\right) +\sin\left(\bar{v}_{\Delta}^{-}\right)\big)\bigg] \\
     2j \sin\bigg[\frac{(m-1)\pi}{2}\bigg(\sin\left(\bar{v}_{\Delta}^{+}\right) -\sin\left(\bar{v}_{\Delta}^{-}\right)\bigg)\bigg] \Bigg|^2
 \end{multline*}
 \begin{equation*}
     = \left(\frac{\bar{v_i}-\bar{v_j}}{\Delta}\right)^2
   \Bigg[\sin\Bigg( \frac{(m-1)\pi}{2} 
   \left( \sin\left(\bar{v}_{\Delta}^{+}\right) -\sin\left(\bar{v}_{\Delta}^{-}\right) \right) \Bigg)\Bigg] ^2.
 \end{equation*}
  Let us define
   $\Delta_{\sin}:=\sin\left(\bar{v}_{\Delta}^{+}\right) -\sin\left(\bar{v}_{\Delta}^{-}\right)$.
  Then, one can write
  \begin{multline}
       \left\|{\textbf{p}_{UL}}^{i}-{\textbf{p}_{UL}}^{j} \right\|^2 
       = \sum_{m=1}^{M} \left|{[{\textbf{p}_{UL}}^i]}_m-{[{\textbf{p}_{UL}}^j]}_m\right|^2 \\
       \approx \left(\frac{\bar{v_i}-\bar{v_j}}{\Delta}\right)^2    
   \sum_{m=1}^{M} \left(\sin\left((m-1)\pi \frac{\Delta_{\sin}}{2}\right)\right) ^2.
  \end{multline}
   Similarly, one can approximate $\|{\textbf{p}_{DL}}^{i}-{\textbf{p}_{DL}}^{j} \|^2$ as 
   \begin{multline}
       \|{\textbf{p}_{DL}}^{i}-{\textbf{p}_{DL}}^{j} \|^2
       = \sum_{m=1}^{M} |{[{\textbf{p}_{DL}}^i]}_m-{[{\textbf{p}_{DL}}^j]}_m|^2 \\
   \approx \left(\frac{\bar{v_i}-\bar{v_j}}{\Delta}\right)^2   
   \sum_{m=1}^{M} \left(\sin\left(f_R (m-1)\pi \frac{\Delta_{\sin}}{2}\right)\right) ^2
   \end{multline}
which yields
   \begin{equation} \label{mult}
      \frac{\left\|{\textbf{p}_{DL}}^{i}-{\textbf{p}_{DL}}^{j} \right\|^2}{\left\|{\textbf{p}_{UL}}^{i}-{\textbf{p}_{UL}}^{j} \right\|^2} 
      \approx  \frac{ \sum_{m=1}^{M} (\sin(f_R (m-1)\pi \frac{\Delta_{\sin}}{2})) ^2}{\sum_{m=1}^{M} (\sin((m-1)\pi \frac{\Delta_{\sin}}{2})) ^2}.
     \end{equation} 
 Let us denote the sine ratio as 
\[R_{\sin} :=\frac{ \sum_{m=1}^{M} (\sin(f_R (m-1)\pi \frac{\Delta_{\sin}}{2})) ^2}{\sum_{m=1}^{M} (\sin((m-1)\pi \frac{\Delta_{\sin}}{2})) ^2}. \]
The constant $K$ introduced in Theorem \ref{lemma-2} can then be defined as the maximum value that $R_{\sin}$ can take.
\end{proof}

\section{\textbf{Proof of Theorem ~\ref{thm:some-theorem}}} \label{Appendix_theorem1}
\begin{proof}[\unskip\nopunct] 
The norm of the difference between an arbitrary test point in the DL CCM dataset and its estimate obtained by the mapping of its UL counterpart via the interpolation function $f\left(.\right)$ can be bounded as 
\begin{multline*}
    \left\| f({\textbf{r}_{UL}}^{test})-{{\textbf{r}}_{DL}}^{test} \right\| = \\
    \Bigg\| f({\textbf{r}_{UL}}^{test})-\frac{1}{|A^{UL}|} \sum_{{{\textbf{r}_{UL}}^{i}} \in A^{UL}}
 f({\textbf{r}_{UL}}^{i}) 
 +\frac{1}{|A^{UL}|} \sum_{{{\textbf{r}_{UL}}^{i}} \in A^{UL}}
 f({\textbf{r}_{UL}}^{i}) -{{\textbf{r}}_{DL}}^{test} \Bigg\| 
 \end{multline*}
 \begin{multline*}
 \leq \left\| f({\textbf{r}_{UL}}^{test})-\frac{1}{|A^{UL}|} \sum_{{{\textbf{r}_{UL}}^{i}} \in A^{UL}}
 f({\textbf{r}_{UL}}^{i}) \right\| 
 + \left\| {{\textbf{r}}_{DL}}^{test} - \frac{1}{|A^{UL}|} \sum_{{{\textbf{r}_{UL}}^{i}} \in A^{UL}}
 f({\textbf{r}_{UL}}^{i}) \right\| \\
 =\left\| f({\textbf{r}_{UL}}^{test})-\frac{1}{|A^{UL}|} \sum_{{{\textbf{r}_{UL}}^{i}} \in A^{UL}}
 f({\textbf{r}_{UL}}^{i})\right\| \\
 + \Bigg\| {{\textbf{r}}_{DL}}^{test} - \frac{1}{|A^{UL}|} \sum_{i : {{\textbf{r}_{UL}}^{i}} \in A^{UL}}
 {\textbf{r}_{DL}}^{i}
 + \frac{1}{|A^{UL}|} \sum_{i : {{\textbf{r}_{UL}}^{i}} \in A^{UL}}
 {\textbf{r}_{DL}}^{i} - \frac{1}{|A^{UL}|} \sum_{{{\textbf{r}_{UL}}^{i}} \in A^{UL}}
 f({\textbf{r}_{UL}}^{i})  \Bigg\| 
 \end{multline*}
 \begin{multline*}
 \leq \left\| f({\textbf{r}_{UL}}^{test})-\frac{1}{|A^{UL}|} \sum_{{{\textbf{r}_{UL}}^{i}} \in A^{UL}}
 f({\textbf{r}_{UL}}^{i})\right\| 
 + \left\| {{\textbf{r}}_{DL}}^{test} - \frac{1}{|A^{UL}|} \sum_{i : {{\textbf{r}_{UL}}^{i}} \in A^{UL}}
 {\textbf{r}_{DL}}^{i}  \right\| \\
 + \left\| \frac{1}{|A^{UL}|} \sum_{i : {{\textbf{r}_{UL}}^{i}} \in A^{UL}}
 {\textbf{r}_{DL}}^{i} - \frac{1}{|A^{UL}|} \sum_{{{\textbf{r}_{UL}}^{i}} \in A^{UL}}
 f({\textbf{r}_{UL}}^{i}) \right\| 
 \end{multline*}
 \begin{multline*}
 \leq \left\| f({\textbf{r}_{UL}}^{test})-\frac{1}{|A^{UL}|} \sum_{{{\textbf{r}_{UL}}^{i}} \in A^{UL}}
 f({\textbf{r}_{UL}}^{i}) \right\| 
 + \left\| {{\textbf{r}}_{DL}}^{test} - \frac{1}{|A^{UL}|} \sum_{i : {{\textbf{r}_{UL}}^{i}} \in A^{UL}}
 {\textbf{r}_{DL}}^{i}  \right\| \\
 +  \frac{1}{|A^{UL}|} \sum_{i : {{\textbf{r}_{UL}}^{i}} \in A^{UL}}
 \left\| {\textbf{r}_{DL}}^{i} - f({\textbf{r}_{UL}}^{i}) \right\| .
\end{multline*}

Let us denote $\left\| f({\textbf{r}_{UL}}^{test})-\frac{1}{|A^{UL}|} \sum_{{{\textbf{r}_{UL}}^{i}} \in A^{UL}}
 f({\textbf{r}_{UL}}^{i})\right\| $ as (UB-1) ,\\ $\left\| {{\textbf{r}}_{DL}}^{test} - \frac{1}{|A^{UL}|} \sum_{i : {{\textbf{r}_{UL}}^{i}} \in A^{UL}}
 {\textbf{r}_{DL}}^{i}  \right\| $ as (UB-2) and \\ $\frac{1}{|A^{UL}|} \sum_{i : {{\textbf{r}_{UL}}^{i}} \in A^{UL}}
 \left\| {\textbf{r}_{DL}}^{i} - f({\textbf{r}_{UL}}^{i}) \right\|$ as (UB-3). \\

 (UB-1) can be upper bounded by using Lemma \ref{lemma-1}, which is the adaptation of Lemma 1 in \cite{kaya2020learning} to our study. The proof of Lemma \ref{lemma-1} is presented in \ref{Appendix_lemma2}. \\
\begin{lemma} \label{lemma-1}
Let the training sample set contain at least $N$ training samples $\{{\textbf{r}_{UL}}^{i}\}_{i=1}^{N}$ with ${\textbf{r}_{UL}}^{i} \sim \upsilon$. Assume that the interpolation function $f:\mathbb{R}^{1 \times 2M-1}\rightarrow\mathbb{R}^{1 \times 2M-1}$ is Lipschitz continuous with constant \textit L.  
Let ${{\textbf{r}}_{UL}}^{test}$ be a test sample drawn from $\upsilon$ independently of the training samples. Let $A^{UL}$ be defined as in \eqref{A_UL}.

Then, for any $\epsilon > 0$, for some $\frac{1}{N\eta_{\delta}} \leq a < 1$ and $ \delta > 0$,  with probability at least 
\begin{multline*}
    \left(1-\exp\left(-2N(\left(1-a\right)\eta_{\delta})^2\right)\right) 
\left(1-2\sqrt{2M-1}\exp\left(-\frac{aN\eta_{\delta}\epsilon^2}{2L^2\delta^2}\right)\right),
\end{multline*}
the set $A^{UL}$ contains at least $aN\eta_{\delta}$ samples and the distance between the embedding of $\textbf{r}_{UL}^{test}$ and the sample mean of the embeddings of its neighboring training samples is bounded as
\begin{multline} \label{UB-1_bound}
\left\| f({\textbf{r}_{UL}}^{test})-\frac{1}{|A^{UL}|} \sum_{{{\textbf{r}_{UL}}^{i}} \in A^{UL}} f({\textbf{r}_{UL}}^{i})\right\| 
\leq L\delta+\sqrt{2M-1}\epsilon. \\
\end{multline}
\end{lemma}

Next, (UB-2) can be bounded by using Theorem \ref{lemma-2} as
\begin{multline} \label{UB-2_bound}
 \left\| {{\textbf{r}}_{DL}}^{test}-\frac{1}{|A^{UL}|}\sum_{{{\textbf{r}_{UL}}^{i}} \in A^{UL}}
 {\textbf{r}_{DL}}^{i}  \right\| 
 = \left\| \frac{1}{|A^{UL}|} \sum_{i : {{\textbf{r}_{UL}}^{i}} \in A^{UL}}
 ({{\textbf{r}}_{DL}}^{test} -{\textbf{r}_{DL}}^{i})  \right\| \\
\leq  \frac{1}{|A^{UL}|} \sum_{i : {{\textbf{r}_{UL}}^{i}} \in A^{UL}} \left\| {{\textbf{r}}_{DL}}^{test} -{\textbf{r}_{DL}}^{i}  \right\| 
\leq  \frac{1}{|A^{UL}|} \sum_{{{\textbf{r}_{UL}}^{i}} \in A^{UL}} K \left\| {{\textbf{r}}_{UL}}^{test} -{\textbf{r}_{UL}}^{i}  \right\| \\
\leq  \frac{1}{|A^{UL}|}|A^{UL}|K\delta = K\delta,   
\end{multline}
for some constant $K>0$.

Finally, (UB-3) is the average training error of the points in $A^{UL}$. Thus, upper bounding  (UB-1) and (UB-2) as in \eqref{UB-1_bound} and \eqref{UB-2_bound} respectively, the difference between the test error of any point and the average training error of its neighboring training points can be upper bounded as given in Theorem \ref{thm:some-theorem}. 
\end{proof}

\section{\textbf{Proof of Lemma~\ref{lemma-1}}} \label{Appendix_lemma2}

\begin{proof} [\unskip\nopunct] 
A training sample ${\textbf{r}_{UL}}^{i}$ drawn independently from ${\textbf{r}_{UL}}^{test}$ lies in the $\delta$-neighborhood of ${\textbf{r}_{UL}}^{test}$ with probability 
\[ P\left({\textbf{r}_{UL}}^{i} \in B_{\delta}\left({\textbf{r}_{UL}}^{test}\right)\right) =  \upsilon \left(B_{\delta}\left({\textbf{r}_{UL}}^{test}\right)\right) \geq \eta_{\delta}. \] 
From \cite{kaya2020learning} and the references therein, one can show that 
\[ P\left(\left|A^{UL}\right| \geq Q \right) \geq 1-\exp\left(-\frac{2\left(N\eta_{\delta}-Q\right)^2}{N}\right),\] 
for $1 \leq Q < N\eta_{\delta}$. 
Assuming that $\left|A^{UL}\right| \geq Q$, from \cite{kaya2020learning} and the references therein, one can show that, with probability at least
\begin{multline*}
    1-2\sqrt{2M-1}\exp\left(-\frac{\left|A^{UL}\right|\epsilon^2}{2L^2\delta^2}\right) 
    \geq 1-2\sqrt{2M-1}\exp\left(-\frac{Q\epsilon^2}{2L^2\delta^2}\right),
\end{multline*}
the distance between the embedding of ${\textbf{r}_{UL}}^{test}$ and the sample average of the embeddings of training samples lying inside the $\delta$-neighborhood of ${\textbf{r}_{UL}}^{test}$ is bounded as
\begin{multline} \label{eqn}
\left \| f({\textbf{r}_{UL}}^{test})-\frac{1}{\left|A^{UL}\right|} \sum_{{{\textbf{r}_{UL}}^{i}} \in A^{UL}}
 f({\textbf{r}_{UL}}^{i})\right\|
 \leq L\delta+\sqrt{2M-1}\epsilon.    
\end{multline}
 
 Let $B_1$ be the event that the inequality in \eqref{eqn} holds.
 Combining the probability expressions above,
 \begin{multline}
 P\left((\left|A^{UL}\right| \geq Q ) \cap B_1\right) 
= 
P\left(\left|A^{UL}\right| \geq Q \right) \ P\left(B_1\ \big| \ (\left|A^{UL}\right| \geq Q)\right) \\
 \geq \left(1-\exp\left(-\frac{2(N\eta_{\delta}-Q)^2}{N}\right)\right)
 \left(1-2\sqrt{2M-1}\exp\left(-\frac{Q\epsilon^2}{2L^2\delta^2}\right)\right). 
\end{multline}
Thus, we obtain that with probability at least 
\begin{multline*}
\left(1-\exp\left(-\frac{2(N\eta_{\delta}-Q)^2}{N}\right)\right) 
\left(1-2\sqrt{2M-1}\exp\left(-\frac{Q\epsilon^2}{2L^2\delta^2}\right)\right),
\end{multline*}
$\left|A^{UL}\right| \geq Q$ and $B_1$ occurs.
Setting $Q=aN\eta_{\delta}$ for $0<a<1$, one can reach the statement given in Lemma \ref{lemma-1}.
\end{proof}

\section{\textbf{Numerical Analysis About the Constant $K$:}} \label{Appendix_K_constant}

Let $C := \cos\left(\frac{\bar{v_i}+\bar{v_j}}{2}\right) $ and $b := C \sin\left(\Delta\right) $. Then, $\Delta_{sin}$
   can be written as 
   \begin{multline} \label{eqn1}
        \Delta_{\sin}=\sin\left(\frac{\bar{v_i}+\bar{v_j}}{2}+\Delta\right)-\sin\left(\frac{\bar{v_i}+\bar{v_j}}{2}-\Delta\right)\\
        = 2 \cos\left(\frac{\bar{v_i}+\bar{v_j}}{2}\right) \sin\left(\Delta\right) =2b.
   \end{multline}

Since $-1 \leq C \leq 1$, we have $- \sin\left(\Delta\right) \leq b \leq \sin\left(\Delta\right)$. Since $\sin^2(\cdot)$ is an even function, it is enough to examine only the positive side of the interval, i.e., $0 \leq b \leq \sin\left(\Delta\right)$. We evaluate the constant $K$ for different $\Delta$ values  (hence, different maximum values of $b$) by investigating the values of the number $M$ of base station antennas  within the range  $2 \leq M \leq 1000$.  Table \ref{fr_delta_table} reports the values that $K$ takes for different $\Delta$ values, where we set $f_R=1.0974$ as in our communication scenario.

\begin{table}[!h]
\caption{$K$ Values for $f_R=1.0974$ and for Different $\Delta$ Values}
\label{fr_delta_table}
\centering
\begin{tabular} { |p{3.5cm}|p{3.5cm}|  } 
 \hline
 $\Delta (\degree)$ & Corresponding $K$ Value \\
 \hline
 5  & 1.0974\\
 \hline
10  & 1.0974\\
 \hline
 15  & 1.0974\\
 \hline
 35  & 1.0974\\
 \hline
 45  & 1.1317\\
 \hline
 60  & 1.1893\\
 \hline
\end{tabular}
\end{table}

\end{document}